%% file: main.tex
\documentclass[lettersize,journal]{IEEEtran}
\usepackage{amsmath,amsfonts}
\usepackage{algorithmic}
\usepackage{algorithm}
\usepackage{array}
\usepackage[caption=false,font=normalsize,labelfont=sf,textfont=sf]{subfig}
\usepackage{textcomp}
\usepackage{stfloats}
\usepackage{url}
\usepackage{verbatim}
\usepackage{graphicx}
\usepackage{cite}
\usepackage{amsmath}
\usepackage{xcolor}
\usepackage{booktabs}
\usepackage{multirow}  
\usepackage{array}  
\usepackage{makecell} 
\usepackage{multicol}

\def\method{OledFL}

\begin{document}

\title{\method{}: Unleashing the Potential of Decentralized Federated Learning via Opposite Lookahead Enhancement}

\author{Qinglun Li, Miao Zhang, Mengzhu Wang, Quanjun Yin and Li Shen
\thanks{\IEEEcompsocthanksitem Qinglun Li is with the National University of Defense Technology, \texttt{liqinglun@nudt.edu.cn}.
\IEEEcompsocthanksitem Miao Zhang is with the National University of Defense Technology, \texttt{zhangmiao15@nudt.edu.cn}.
\IEEEcompsocthanksitem Mengzhu Wang is with the Hebei University of Technology,  Tianjin, 300000, China, \texttt{dreamkily@gmail.com}. 
\IEEEcompsocthanksitem Quanjun Yin is with the National University of Defense Technology, \texttt{yin\_quanjun@163.com}. 
\IEEEcompsocthanksitem Li Shen is with School of Cyber Science and Technology, Shenzhen Campus of Sun Yat-sen University, Shenzhen 518107, China. \texttt{mathshenli@gmail.com}. 
\IEEEcompsocthanksitem  Manuscript received April XX, XXXX; revised August XX, XXXX.}}


\markboth{Journal of \LaTeX\ Class Files,~Vol.~14, No.~8, August~2021}%
{Shell \MakeLowercase{\textit{et al.}}: A Sample Article Using IEEEtran.cls for IEEE Journals}


\maketitle

\begin{abstract}
Decentralized Federated Learning (DFL) surpasses Centralized Federated Learning (CFL) in terms of faster training, privacy preservation, and light communication, making it a promising alternative in the field of federated learning. However, DFL still exhibits significant disparities with CFL in terms of generalization ability such as rarely theoretical understanding and degraded empirical performance due to severe inconsistency. In this paper, we enhance the consistency of DFL by developing an opposite lookahead enhancement technique (Ole), yielding  \method{} to optimize the initialization of each client in each communication round, thus significantly improving both the generalization and convergence speed. Moreover, we rigorously establish its convergence rate in non-convex setting and characterize its generalization bound through uniform stability, which provides concrete reasons why \method{} can achieve both the fast convergence speed and high generalization ability. Extensive experiments conducted on the CIFAR10 and CIFAR100 datasets with Dirichlet and Pathological distributions illustrate that our \method{} can achieve up to 5\% performance improvement and 8$\times$ speedup, compared to the most popular DFedAvg optimizer in DFL.
\end{abstract}


\input{tex/1.induction}

\input{tex/2.related_work}
\input{tex/3.methodology}
\input{tex/4.convergence}

\input{tex/5.experiment}

\input{tex/6.conclusion}

\bibliographystyle{IEEEtran}
\bibliography{main.bib}

\onecolumn

\input{tex/8.proof}


 





\end{document}

%% file: tex/1.induction.tex
\newtheorem{theorem}{Theorem}
\newtheorem{proposition}{Proposition}
\newtheorem{lemma}{Lemma}
\newtheorem{corollary}{Corollary}
\newtheorem{definition}{Definition}
\newtheorem{assumption}{Assumption}
\newtheorem{remark}{Remark}
\newtheorem{proof}{Proof}

\section{Introduction}

Federated Learning (FL) is a novel distributed machine learning paradigm that ensures privacy protection \cite{mcmahan2017communication,Gu_Xu_Huo_Deng_Huang_2022,Zhou_Wang_Guo_Gong_Zheng_2019}. It enables multiple participants to collaborate on training models without sharing their raw data. Currently, most research efforts \cite{zhou2023federated,sun2023fedspeed,Li_Sahu_Zaheer_Sanjabi_Talwalkar_Smith_2018, acar2021federated,zhang2021fedpd,dai2023fedgamma} have focused on Centralized Federated Learning (CFL). However, the presence of a central server in CFL introduces challenges such as communication burden, single point of failure \cite{chen2023enhancing}, and privacy breaches \cite{gabrielli2023survey}. In contrast, Distributed Federated Learning (DFL) offers improved privacy protection \cite{cyffers2022privacy}, faster model training \cite{lian2017can}, and robustness to slow client devices \cite{neglia2019role}. Therefore, DFL has emerged as a popular alternative solution \cite{chen2023enhancing, lian2017can}.

However, there still exists a significant performance gap between CFL and existing DFL methods, which we attribute to the decentralized communication approach of DFL. The lack of central server coordination results in increasing inconsistency between the model parameters $\mathbf{x}_i$ of client $i$ and $\mathbf{x}_j$ of client $j$ as the communication rounds increase, the severe inconsistency leads to a significant gap between the final output of the model $\bar{\mathbf{x}}$ and the global optimum $\mathbf{x}^*$, leading to performance disparities compared to CFL. Although existing methods such as DFedAvgM \cite{Sun2022Decentralized} and DFedSAM \cite{shi2023improving} have improved the performance of DFL algorithms by introducing new local optimizers to accelerate convergence and address heterogeneous data overfitting, they have not tackled the issue from the critical perspective of communication discrepancies between DFL and CFL, resulting in a persistent performance gap between DFL and CFL. Additionally, existing theoretical analyses of DFL methods are limited to convergence analysis, while the generalization theory is still absent.

In this work, we propose a plug-in method named the opposite Lookahead enhancement (\textbf{Ole}) technique to enhance the consistency of DFL, dubbed \method{} that reduces the performance gap between CFL and DFL. \method{} framework can seamlessly integrate existing DFL optimizers to significantly improve the convergence speed and generalization performance of DFL empirically. We also theoretically prove that \method{} can enhance the convergence speed and reduce generalization error compared to the original DFL methods. Specifically, \method{} performs a retraction operation during the initialization phase of optimizing each client model parameter $\mathbf{x}_{i}^t$ (see Figure \ref{fig:OledFL}). This ensures that each client will not stray too far from $\mathbf{x}_i^t$ during the subsequent optimization process. By adopting this initialization operation for each client before optimization, the mutual consistency among them is naturally strengthened.

Theoretically, we jointly analyze the optimization error and generalization error by introducing excess error. Specifically, we demonstrate that in a non-convex setting, \method{}-SGD (integrating \method{} with DFedAvg) and \method{}-SAM (integrating \method{} with DFedSAM) exhibit an optimization error and convergence rate of $\mathcal{O}(\frac{1}{\sqrt{KT}})$, and the theoretical analysis indicates that ``Ole" can reduce the algorithm's optimization error (Remark \ref{remark:2}). Additionally, through uniform stability analysis, we assess the algorithm's generalization error and find that ``Ole" can significantly reduce generalization error (Remark \ref{remark:gener_2}). For the experiments on CIFAR10\&100 datasets under Dirichlet and Pathological distributions, \method{} consistently demonstrates notable improvements in convergence speed (see Table \ref{ta:CIFAR10-convergence}) and generalization performance (see Table \ref{ta:all_baselines}) over existing DFL methods and outperforms state-of-the-art CFL methods such as FedSAM \cite{qu2022generalized} and SCAFFOLD \cite{scaffold2020}.

In summary, our main contributions are three-fold:
\begin{itemize}  
    \item We propose an opposite Lookahead enhancement (Ole) technique to address the inconsistency issue in DFL, dubbed \method{}. \method{} can seamlessly integrate existing DFL optimizers such as DFedAvg and DFedSAM to significantly improve their convergence speed and generalization performance, thereby reducing the performance gap between CFL and DFL. Furthermore, we provide a comprehensive explanation of the role of Ole, including intuitive (Figure \ref{fig:OledFL}), theoretical (Remark \ref{remark:gener_2}), and experimental interpretations (Section \ref{se:explain ole} \& Figure \ref{fig:comprision}).
    \item We establish a pioneering generalization analysis (Section \ref{section:Theoretical Analysis}) in the field of DFL, which theoretically demonstrates the effectiveness of \method{} from the perspective of optimization error bounds and generalization error bounds.   
    \item We conduct extensive experiments on CIFAR10\&100 datasets under Dirichlet and Pathological distributions, respectively, which demonstrate that \method{} significantly enhances the convergence speed and generalization performance of existing DFL methods. With the assistance of Ole, it can achieve up to 5\% performance improvement and 8$\times$ speedup, compared to the most popular DFedAvg optimizer in DFL. This notably reduces the gap between CFL and DFL (Table \ref{ta:all_baselines} \& Figure \ref{fig:Compared_baselines-cifar10}).
\end{itemize}

%% file: tex/2.related_work.tex
\section{Related work}

\begin{table*}[t]
\caption{the theoretical differences between the baseline methods and the \method{}. Note that \method{} supports generalization analysis.}
\label{ta:difference in theoretical}
\centering
\resizebox{1\textwidth}{!}{%
\begin{tabular}{|l|c|c|c|c|c|}
\hline
\textbf{Method} & \textbf{Non-convex} &\textbf{Convergence Analysis} & \textbf{Generalization Analysis } & \textbf{Multi-local Update } & \textbf{No bounded gradient} \\
\hline
DPSGD \cite{lian2017can,sun2021stability} & $\checkmark$ & $\checkmark$ & $\checkmark$ & $\times$ & $\checkmark$ \\
\hline
DFedAvg \cite{Sun2022Decentralized} & $\checkmark$ & $\checkmark$ & $\times$ & $\checkmark$ & $\times$ \\
\hline
DFedAvgM \cite{Sun2022Decentralized} & $\checkmark$ & $\checkmark$ & $\times$ & $\checkmark$ & $\times$ \\
\hline
DFedSAM \cite{shi2023improving} & $\checkmark$ & $\checkmark$ & $\times$ & $\checkmark$ & $\checkmark$ \\
\hline
\method{} [\textbf{ours}]  & $\checkmark$ & $\checkmark$ & $\checkmark$ & $\checkmark$ & $\checkmark$ \\
\hline
\end{tabular} }
\vspace{-0.3cm}
\end{table*}

Below, we will briefly review the most relevant work to our research, which includes Decentralized Federated Learning (DFL), acceleration techniques in optimization, and theoretical guarantees of DFL.

\textbf{Decentralized Federated Learning (DFL).} To mitigate the communication burden on the server in centralized scenarios, decentralized communication methods distribute the communication load to each node while maintaining overall communication complexity equivalent to that in centralized scenarios \cite{lian2017can}. Additionally, decentralized communication methods afford improved privacy protection compared to CFL \cite{yang2019federated,lalitha2018fully, lalitha2019peer}. DFL has emerged as a promising field of research, recognized as a challenge in various review articles in recent years \cite{beltran2022decentralized, Kairouz2021Advances}. Within DFL, Sun et al. \cite{Sun2022Decentralized} extend the FedAvg algorithm \cite{mcmahan2017communication} to decentralized scenarios and complement it with local momentum acceleration to enhance convergence. Furthermore, Dai et al. \cite{dai2022dispfl} introduce sparse training into DFL to reduce communication and computation costs, while shi et al. \cite{shi2023improving} apply SAM to DFL and enhance the consistency among clients by incorporating Multiple Gossip Steps. These endeavors gradually improve the performance of DFL from different perspectives. However, a significant performance gap still exists compared to CFL. For further related work on DFL, please refer to the survey papers \cite{gabrielli2023survey,yuan2023decentralized,beltran2022decentralized} and their references therein. 

\textbf{Acceleration Techniques for Deep Learning.} In deep learning, momentum and restart are typical acceleration techniques. The momentum techniques focus more on the optimization of the optimizer design, while restart techniques focus more on the selection of initialization points. In Federated Learning (FL), one type of method used in the centralized FL (CFL) domain is the global momentum, such as FedCM \cite{xu2021fedcm} and MimeLite \cite{karimireddy2020mime}, which estimates the global momentum at the server and applies it to each client update, thereby alleviating the problem of client heterogeneity. Another type is the local momentum used in DFL, such as the DFedAvgM algorithm proposed by \cite{Sun2022Decentralized}, which utilizes local momentum to accelerate the convergence process. Restart techniques have nearly negligible computational cost but significantly enhance the effectiveness of the algorithm. Zhou \cite{Zhou2021towards} introduces the Lookahead optimizer, which improves learning stability and reduces the variance of its inner optimizer through a $k$-steps forward, $1$ step back approach. Additionally, lin et al.\cite{lin2015universal} develop a universal framework called Catalyst, which can accelerate first-order optimization methods such as SAG \cite{schmidt2017minimizing}, SAGA \cite{defazio2014saga}, and SVRG \cite{xiao2014proximal}. Subsequently, Trimbach et al.\cite{trimbach2021acceleration} extend Catalyst to the field of distributed learning and used Catalyst to accelerate the DSGD algorithm \cite{koloskova2020unified}.

\textbf{Theoretical Guarantees of DFL.} 
A more comprehensive comparison is presented in Table \ref{ta:difference in theoretical}.
In the aspect of convergence analysis, current DFL works such as DFedAvg \cite{Sun2022Decentralized}, DFedAvgM \cite{Sun2022Decentralized}, and DFedSAM \cite{shi2023improving} have only concentrated on the convergence analysis. DFedSAM, by abandoning the gradient bounded assumption in DFedAvg, has demonstrated an advancement in the convergence analysis of the algorithm. However, in machine learning, algorithms are often evaluated based on their ability to perform well on new, unseen data. This is known as generalization performance, and algorithms that exhibit higher generalization performance are considered to be state-of-the-art. However, in the field of DFL, there is currently a lack of analysis on how well-proposed algorithms generalize to new data. This results in a lack of theoretical support for why these algorithms perform well. 


%% file: tex/3.methodology.tex
\section{Methodology}

In this section, we begin by elucidating the meanings of several notations. We then introduce the problem setup, and then, we propose \method{} and compare it with existing methods to demonstrate the novelty of \method{}. Finally, we will elucidate the relation with Chebyshev Acceleration and provide an effective and intuitive explanation of the \method{}.

\subsection{Notations}

Let $m$ be the total number of clients.  $T$ represents the number of communication rounds. $(\cdot)_{i,k}^{t}$ indicates variable $(\cdot)$ at the $k$-th iteration of the $t$-th round in the $i$-th client. $\mathbf{x}$ denotes the model parameters. 
The communication topology between clients can be represented as graph denoted as $\mathcal{G} = (\mathcal{N}, \mathcal{E}, {\bf W})$, where $\mathcal{N} = \{1, 2, \ldots, m\}$ represents the set of clients, $\mathcal{E} \subseteq \mathcal{N} \times \mathcal{N}$ denotes the links between clients,
$w_{i,j}$ represents the weight of the link between client $j$ and client $i$ and ${ \bf W} = [w_{i,j}] \in [0,1]^{m\times m}$ represents the mixing matrix. The inner product of two vectors is denoted by $\langle\cdot,\cdot\rangle$, and $\Vert \cdot \Vert$ represents the Euclidean norm of a vector. Other symbols will be explained in their respective contexts.


\subsection{Problem Setup}
In this paper, we consider a network of $m$ clients whose objective is to jointly solve the following distributed population risk \(F\) minimization problem:
\begin{equation}\label{finite_sum}
    \small \min_{{\bf x}\in \mathbb{R}^d} F({\bf x}):=\frac{1}{m}\sum_{i=1}^m F_i({\bf x}),~~F_i({\bf x})=\mathbb{E}_{\xi\sim \mathcal{D}_i} F_i({\bf x};\xi)
\end{equation}
where $\mathcal{D}_i$ represents the data distribution in the $i$-th client, which exhibits heterogeneity across clients. Each client's local objective function $F_i({\bf x};\xi)$ is associated with the training data samples $\xi$. We denote $\mathbf{x}_{\mathcal{D}}^\star = \arg \min_{\mathbf{x}}F(\mathbf{x})$ as the optimal value of (\ref{finite_sum}). Unlike CFL, we address (\ref{finite_sum}) by enabling clients to collaborate through a mixing matrix $\mathbf{W}$ in a decentralized manner, leveraging peer-to-peer communication among clients without the need for server coordination. 

Practically, we consider the empirical risk minimization of the non-convex finite-sum problem in DFL as:
\begin{equation}\label{finite_sum_empirical}
    \small \min_{{\bf x}\in \mathbb{R}^d} f({\bf x}):=\frac{1}{m}\sum_{i=1}^m f_i({\bf x}),~~f_i({\bf x})=\frac{1}{S_i}\sum_{z_j\in\mathcal{S}_i}f_i({\bf x};z_j)
\end{equation}
where each client stores a private dataset \(S_i = \{z_j\}\), with \(z_j\) drawn from an unknown distribution \(\mathcal{D}_i\). We denote $\mathbf{x}^\star = \arg \min_{\mathbf{x}}f(\mathbf{x})$ as the optimal value of problem (\ref{finite_sum_empirical}).

\subsection{\method{} Algorithm}\label{sec:oledfl alg}

\begin{figure}[t]
    \vspace{-0.5cm}
    \centering
    \subfloat{%
        \includegraphics[width=0.5\textwidth]{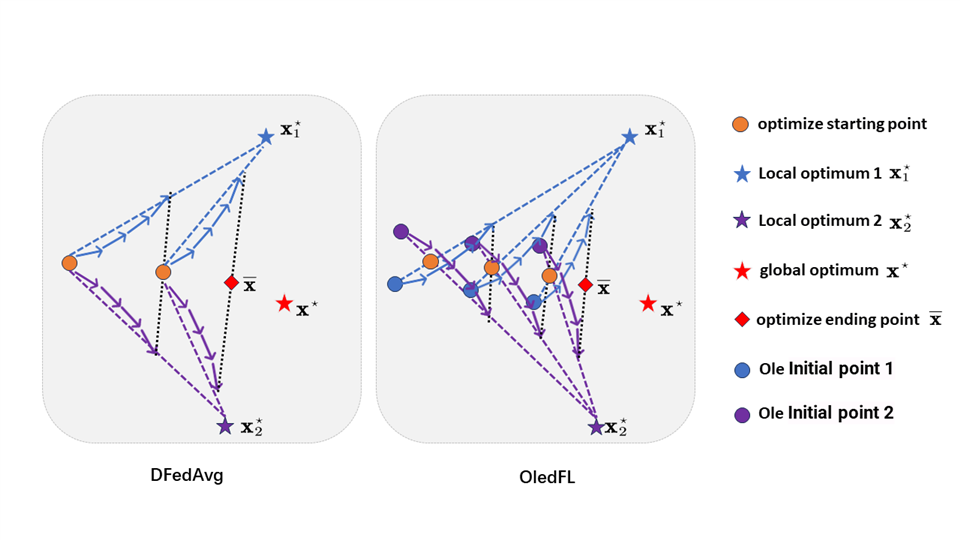}}
    \vspace{-0.3cm}
    \caption{Simulate the optimization process diagrams for two clients under the DFedAvg and \method{} algorithms. Due to the presence of \( \mathbf{x}_i^t - \mathbf{x}_{i,K}^{t-1} \) (\textbf{Ole}), where $\mathbf{x}_{i,K}^{t-1} \approx \mathbf{x}_{i}^*$, then the Ole initial point in \method{} represents taking a step back along the direction from the optimize starting point to the local optimum of the client. Furthermore, from the length of the dashed lines in the figure, it is evident that Ole significantly reduces the inconsistency during the optimization process.}
    \label{fig:OledFL}
\end{figure}

The most popular decentralized FL optimizers for solving the problem (\ref{finite_sum_empirical}) are  DFedSAM \cite{shi2023improving}
and DFedAvg \cite{Sun2022Decentralized}. However, the decentralized approach, lacking central server coordination, leads to increased client inconsistency. To enhance consistency, we design the opposite lookahead enhanced (\textbf{Ole}) initialization method before performing local optimization at each client:
\begin{equation}\label{eq:ole}
    \mathbf{x}_{i,0}^{t}=\mathbf{x}_i^{t} + \beta\underbrace{ (\mathbf{x}_i^t - \mathbf{x}_{i,K}^{t-1})}_{\mathbf{Ole}}
\end{equation}
where $\mathbf{x}_i^{t} = \sum_j w_{i,j} \mathbf{x}_{j,K}^{t-1}$ is the aggregated model. Intuitively, at each initialization, clients obtain the opposite direction of the most recent iteration through Ole. This ensures that the values obtained in the subsequent optimization, $\mathbf{x}_{i,K}^{t}$, do not deviate too far from the initial value $\mathbf{x}_i^{t}$, thereby strengthening the consistency among clients. A more intuitive explanation of the effect of Ole is shown in Figure \ref{fig:OledFL}. The complete algorithm is presented in Algorithm \ref{alg:OledFL} \footnote{Here we provide the default form of \method{} that is composed with SAM. In the experiments, we will instantiate \method{} as \method{}-SGD (by setting $\lambda = 0$) and \method{}-SAM if necessary.}.

\begin{algorithm}[t]
\small
\renewcommand{\algorithmicrequire}{\textbf{Input:}}
\renewcommand{\algorithmicensure}{\textbf{Output:}}
\caption{ \method{} Algorithm}
    \begin{algorithmic}[1]\label{alg:OledFL}
        \REQUIRE Initialize $\eta,\lambda > 0$  and $\mathbf{x}_{i}^{0} \!=\! \mathbf{x}_{i,K}^{-1} \!=\! \mathbf{x}^0 \!\in\! \mathbb{R}^d$ for all nodes.\\
        \ENSURE model average parameters $\bar{\mathbf{x}}^{t}$.
        \FOR{$t = 0, 1, 2, \cdots, T-1$}
        \FOR{client $i$ in parallel}
        \STATE set $\mathbf{x}_{i,0}^{t}=\mathbf{x}_i^{t} + \beta (\mathbf{x}_i^t - \mathbf{x}_{i,K}^{t-1})$
        \FOR{$k = 0, 1, 2, \cdots, K-1$}
        \STATE sample a minibatch $\varepsilon_{i,k}^{t}$ and do
        \STATE estimate stochastic gradient: $\mathbf{g}_{i,k,1}^{t}=\nabla f_{i}(\mathbf{x}_{i,k}^{t};\varepsilon_{i,k}^{t})$
        \STATE update extra step: $\Breve{\mathbf{x}}_{i,k}^{t}=\mathbf{x}_{i,k}^{t}+\lambda\frac{\mathbf{g}_{i,k,1}^{t}}{\Vert\mathbf{g}_{i,k,1}\Vert}$
        \STATE estimate stochastic gradient: $\mathbf{g}_{i,k}^{t}=\nabla f_{i}(\Breve{\mathbf{x}}_{i,k}^{t};\varepsilon_{i,k}^{t})$
        \STATE perform SGD step: $\mathbf{x}_{i,k+1}^{t}=\mathbf{x}_{i,k}^{t}-\eta\mathbf{g}_{i,k}^t$
        \ENDFOR
        \STATE $\mathbf{z}_i^t = \mathbf{x}_{i,K}^t$
        \STATE Mix the received model $\mathbf{z}_{j}^{t}$ with mixing matrix ${\bf W}$ (Refer to Definition \ref{def:mixing matrix}): $\mathbf{x}_i^{t+1} = \sum_j w_{i,j} \mathbf{z}_j^t$ 
        \ENDFOR
        \ENDFOR
    \end{algorithmic}
\end{algorithm}

\textbf{Discuss on Lookhead Optimizer.}
Zhang et al. \cite{zhang2019lookahead} propose the lookahead optimizer with the initial optimization value denoted as $\mathbf{x}_0^t$, whose core iterative form is given by:
\begin{equation}
    \mathbf{x}_0^t = \phi^{t-1} + \beta (\mathbf{x}_{K}^{t-1} - \phi^{t-1})
\end{equation}
It optimizes the sequence of ``fast weights" $\mathbf{x}$ through an internal loop $K$ times and then utilizes these fast weights to determine the initial search direction of the ``slow weights" $\phi$, which reduces variance and facilitates rapid convergence of lookahead in practice. When we omit the subscript $i$ in (3) and set $\phi^{t-1} =  \sum_j w_{i,j} \mathbf{x}_{j,K}^{t-1}$ in (4), the Ole component in (3) is exactly opposite to the symbol in (4), which is the origin of the name ``Opposite lookahead enhancement".

\textbf{Discuss on Catalyst.} In distributed optimization, lin et al.\cite{lin2015universal} propose a generic acceleration scheme for a large class of optimization methods with the initial optimization value denoted as $\mathbf{x}_0^t$, whose core iterative form as follows:
\begin{equation}\label{eq:catalyst}
    \mathbf{x}_0^t = \mathbf{x}^{t} + \beta^t (\mathbf{x}^{t} - \mathbf{x}^{t-1})
\end{equation}
Through (5), significant acceleration in the convergence of algorithms such as SAG \cite{schmidt2017minimizing}, SAGA \cite{defazio2014saga}, and SVRG \cite{xiao2014proximal} can be achieved. By comparing (\ref{eq:catalyst}) with (\ref{eq:ole}), it is evident that the initialization coefficient $\beta$ in (\ref{eq:ole}) is a simpler fixed value. When considering the subscript $i$ in (\ref{eq:catalyst}) \cite{trimbach2021acceleration}, the subtraction in Ole is performed using the client's most recently obtained parameter value $\mathbf{x}_{i,K}^{t-1}$ rather than the value after the previous communication $\mathbf{x}_{i}^{t-1}$. 
This difference leads to a fundamental distinction between catalyst and ole, where ole initializes the point in the opposite direction of the current point and the local optimal value point, as shown in Figure \ref{fig:OledFL} (\method{}), while catalyst, similar to the momentum method, initializes the point in the direction of $\mathbf{x}^{t} - \mathbf{x}^{t-1}$ from the current point.


\textbf{Discussion of the relation with Chebyshev Acceleration (CA):
} Chebyshev Acceleration has been widely utilized to expedite the attainment of consensus among networked agents during the distributed optimization process \cite{Scaman2017}. When CA is integrated with stochastic gradient accumulation employing a large batch size at each iteration, it can ensure optimal convergence \cite{Lu2020,Yuan2022}. A key idea of CA involves transforming the mixing matrix from $\bf W$ to $\tilde{\bf{W}}$. Here, $\widetilde{\bf W}$ is expressed as:
\begin{equation*}
    \widetilde{\bf W}:=\begin{pmatrix}(1+\eta)\bf W&-\eta \bf I\\\bf I&0\end{pmatrix}
\end{equation*}
In simple terms, CA transforms the communication topology $\bf W$ into $(1+\eta)\bf W - \eta \bf I$ in the communication period, where $\eta < 1$. It can be proven that $\widetilde{\bf W}$ facilitates faster achievement of consensus among multiple clients compared to $\bf W$ \cite{huang2024accelerated, Song2021}. The update form of Ole is equivalent to aggregation using the communication matrix $(1+\beta)\bf W - \beta \bf I$, demonstrating a connection between Ole and the core idea of CA. With the relationship to CA, we will constrain $\beta < 1$ in Algorithm \ref{alg:OledFL}. Figure 1 shows that Ole enhances consistency, and the relationship between Ole and CA can explain that the acceleration effect of CA originates from the enhancement of consistency among clients.

%% file: tex/4.convergence.tex
\section{Theoretical Analysis}\label{section:Theoretical Analysis}

In this section, we begin by presenting the theoretical analysis, which offers a comprehensive examination of the combined performance in optimization and generalization. Then, we introduce the necessary assumptions utilized in our proofs. At last, we outline the main theorems, encompassing both optimization and generalization, respectively.

\subsection{Excess Risk Error}

In existing literature on DFL, the most of analyses focus on the studies from the onefold perspective of convergence but ignore learning its impact on generality \cite{shi2023improving,Sun2022Decentralized,li2023dfedadmm}. Additionally, some studies in distributed learning exclusively analyze the algorithm's generalization while overlooking the effect on convergence \cite{Sun2022Stability,Zhu2023Stability,zhu2022topology}. To offer a more comprehensive examination of the joint performance of both optimization and generalization in FL, we introduce the well-known concept of excess risk in our analysis.


Firstly, we define \(\bar{\mathbf{x}}^T\) as the final model generated by the \method{} method after \(T\) communication rounds, where \(\bar{\mathbf{x}}^T = \frac{1}{m}\sum_{i=1}^m \mathbf{x}_i^T\). In comparison with \(f(\bar{\mathbf{x}}^T)\), our primary focus is on the efficiency of \(F(\bar{\mathbf{x}}^T)\) which corresponds to its generalization performance and test accuracy. Consequently, we analyze \(\mathbb{E}[F(\bar{\mathbf{x}}^T)]\) based on the excess risk \(\mathcal{E}_E\) as:
\begin{equation}
\begin{aligned}
    \mathcal{E}_E&=\mathbb{E}[F(\bar{\mathbf{x}}^T)]-\mathbb{E}[f(\mathbf{x}^\star)] \\
    & =\underbrace{\mathbb{E}[F(\bar{\mathbf{x}}^T)-f(\bar{\mathbf{x}}^T)]}_{\mathcal{E}_G:\text{ generalization error}} + \underbrace { \mathbb{E}[f(\bar{\mathbf{x}}^T)-f(\mathbf{x}^\star)]}_{\mathcal{E}_O:\text{ optimization error}}
\end{aligned}
\end{equation}

Generally,  $\mathbb{E}[f(\mathbf{x}^\star)]$ is expected to be very small and even to zero if the model could fit the dataset. Thus $\mathcal{E}_E$ could be considered as the joint efficiency of the generated model \(\bar{\mathbf{x}}^T\). Thereinto, $\mathcal{E}_G$ means the different performance of \(\bar{\mathbf{x}}^T\) between the training dataset and the test dataset, and $\mathcal{E}_O$ means the similarity between \(\bar{\mathbf{x}}^T\) and optimization optimum \(\mathbf{x}^\star\). From the perspective of the excess risk, $\mathcal{E}_E$ approximates our focus \(\mathbb{E}[F(\bar{\mathbf{x}}^T)]\). It is worth noting that, in non-convex settings, $\mathcal{E}_E$ is measured via gradient residual $\mathbb{E}\| \nabla f(\bar{\mathbf{x}}^T) \|^2$ rather than $f(\bar{\mathbf{x}}^T) - f(\mathbf{x}^\star)$.
We will investigate the optimization and generalization performance respectively in the following subsections.


\subsection{Definition And Assumptions}

Below, we introduce the definition of the mixing matrix and several assumptions utilized in our analysis. 

\begin{definition}\label{def:mixing matrix}
    (Gossip/Mixing matrix). [Definition 1, \cite{Sun2022Decentralized}] The gossip matrix ${ \bf W} = [w_{i,j}] \in [0,1]^{m\times m}$  is assumed to have the following properties:
    (i) \textbf{(Graph)} If $i\neq j$ and $(i,j) \notin {\cal V}$, then $w_{i,j} =0$, otherwise, $w_{i,j} >0$;
    (ii) \textbf{(Symmetry)} ${\bf W} = {\bf W}^{\top}$;
    (iii) \textbf{(Null space property)} $\mathrm{null} \{{\bf I}-{\bf W}\} = \mathrm{span}\{\bf 1\}$;
    (iv) \textbf{(Spectral property)} ${\bf I} \succeq {\bf W} \succ -{\bf I}$. 
    Under these properties, the eigenvalues of $\bf{\bf W}$ satisfy $1=|\psi_1({\bf W)})|> |\psi_2({\bf W)})| \ge \dots \ge |\psi_m({\bf W)})|$. Furthermore, we define $\psi:=\max\{|\psi_2({\bf W)}|,|\psi_m({\bf W)})|\}$ and $1-\psi \in (0,1]$ as the spectral gap of $\bf W$.
\end{definition}

\begin{assumption}\label{as:smoothness}
    (\textbf{L-Smoothness}) \textit{The non-convex function $f_{i}$ satisfies the smoothness property for all $i\in[m]$, i.e., $\Vert\nabla f_{i}(\mathbf{x})-\nabla f_{i}(\mathbf{y})\Vert\leq L\Vert\mathbf{x}-\mathbf{y}\Vert$, for all $\mathbf{x},\mathbf{y}\in\mathbb{R}^{d}$.}
\end{assumption}

\begin{assumption}\label{as:bounded_stochastic_gradient}
(\textbf{Bounded Variance}) The stochastic gradient $\mathbf{g}_{i,k}^{t}=\nabla f_{i}(\mathbf{x}_{i,k}^{t}, \varepsilon_{i,k}^{t})$ with the randomly sampled data $\varepsilon_{i,k}^{t}$ on the local client $i$ is unbiased and with bounded variance, i.e., $\mathbb{E}[\mathbf{g}_{i,k}^{t}]=\nabla f_{i}(\mathbf{x}_{i,k}^{t})$ and $\mathbb{E}\Vert \mathbf{g}_{i,k}^{t} - \nabla f_{i}(\mathbf{x}_{i,k}^{t})\Vert^{2} \leq \sigma_{l}^{2}$, for all $\mathbf{x}_{i,k}^{t}\in\mathbb{R}^{d}$.
 \end{assumption}

\begin{assumption}\label{as:bounded_heterogeneity}
(\textbf{Bounded Heterogeneity}) 
For all $\mathbf{x}\in\mathbb{R}^{d}$.the heterogeneous similarity is bounded on the gradient norm as $\mathbb{E}\|\nabla f_i(w)\|^2\leq G^2+B^2\mathbb{E}\|\nabla f(w)\|^2$, where $G \geq 0$ and $B \geq 1$ are two  constants.
\end{assumption}

\begin{assumption}\label{as:L_G-lip}
    (Lipschitz Continuity). The global function fsatisfies the $L_G$-Lipschitz property, i.e. for $\forall \mathbf{x}_1, \mathbf{x}_2$, $\|f(\mathbf{x}_1) - f(\mathbf{x}_2)\| \leq L_G\| \mathbf{x}_1 - \mathbf{x}_2 \|.$
\end{assumption}

Definition \ref{def:mixing matrix} is commonly used to describe the communication topology in DFL, where it can also be viewed as a Markov transition matrix. The term \(1-\psi\) measures the speed of convergence to the equilibrium state. Assumptions \ref{as:smoothness}-\ref{as:bounded_heterogeneity} are mild and commonly used to analyze the non-convex objective DFL \cite{Sun2022Decentralized, shi2023improving}. Assumption \ref{as:L_G-lip} is utilized to bound the uniform stability for the non-convex objective \cite{Sun2022Stability, sun2023understanding, Hardt2016train}. It is important to note that when describing the algorithm's convergence speed, we only use assumptions \ref{as:smoothness}-\ref{as:bounded_heterogeneity}; whereas in describing the algorithm's generalization bound, we utilize assumptions \ref{as:smoothness}, \ref{as:bounded_stochastic_gradient} and \ref{as:L_G-lip}. This is because assumption \ref{as:L_G-lip} implies bounded gradients, i.e., \(\|\nabla f_i(\mathbf{x})\| \leq L_G\), while assumption \ref{as:bounded_heterogeneity} is more general than the bounded gradient assumption.

\subsection{Optimization Analysis}
In this part, we present the optimization error and convergence rate of \method{} under the assumptions \ref{as:smoothness}-\ref{as:bounded_heterogeneity}. All the proofs can be found in the Appendix \ref{appendix:optimization error}.

\begin{theorem}\label{the:opt_error}
Under Assumption \ref{as:smoothness} - \ref{as:bounded_heterogeneity}, let the learning rate satisfy $\eta \leq  \frac{1}{K^{3/2}LB}$ where $K \geq 2$, let the Ole parameter $\beta \leq \min\{\frac{\sqrt{10}(1-\psi)}{40}, \frac{\sqrt{5}}{30} \}$, and after training $T$ rounds, the averaged model parameters generated by our proposed algorithm satisfies: 
\begin{equation*}
    \begin{aligned}
        &\frac{1}{T}\sum_{t=0}^{T-1}\mathbb{E}\| \nabla f(\bar{\mathbf{x}}^t) \|^2 
        \leq  \frac{\mathbb{E}[f(\bar{\mathbf{x}}^0) - f(\bar{\mathbf{x}}^{T})]}{\kappa \eta K T} + \zeta(\eta,L,\psi)\sigma_l^2 \\
        & + \alpha(\eta,K,L,\psi)G^2 + \phi(K,\eta ,L)\lambda^2 - \chi(L,\psi, \Delta^T, T)\beta^2
    \end{aligned}
\end{equation*}
where $\kappa \in (0,1)$ is a constant and   
$\alpha(\eta,K,L,\psi) = \frac{9}{\kappa}\eta^2 K^2 L^2\left( 1 + \frac{24}{(1-\psi)^2}\right)$,   
$\phi(K,\eta ,L) = \frac{\eta L}{\kappa}(1 + 9K^2\eta L)$,  
$\chi(L,\psi, \Delta^T, T) = \left(3 + \frac{72}{(1-\psi)^2}\right)\frac{L^2\Delta^{T}}{\kappa(1-\gamma)T}$, which the consistency term $\Delta^{T} = \frac1m\sum_{i=1}^m\mathbb{E}\|\mathbf{x}_{i,K}^{T-1}-\mathbf{x}_i^{T}\|^{2}$   
and $\zeta(\eta,L,\psi) = \frac{1}{\kappa} \eta L\left( 2 + \frac{36}{(1-\psi)^2}\right)$.


Further, by selecting the proper learning rate $\eta = \mathcal{O}(\frac{1}{\sqrt{KT}})$ and let $D = f(\bar{\mathbf{x}}^0) - f(\bar{\mathbf{x}}^*)$ as the initialization bias, thenthe averaged model parameters $\bar{\mathbf{x}}^t$ satisfies:
\begin{equation*}
\begin{aligned}
    &\frac{1}{T}\sum_{t=0}^{T-1}\mathbb{E}\| \nabla f(\bar{\mathbf{x}}^t) \|^2 = \mathcal{O}\left(\frac{D}{\sqrt{KT}} + \frac{KL^2}{T(1-\psi)^2}G^2\right. \\
    & \qquad \qquad \qquad + \left.\frac{L}{\sqrt{T}K(1-\psi)^2}\sigma_l^2 + \frac{L}{\sqrt{T}K}\lambda^2\right)
\end{aligned}
\end{equation*}
\end{theorem}

\begin{remark}\label{remark:1}
    When setting \(\lambda = 0\) in Theorem 1, we can obtain the optimization error and convergence rate of \method{} with local SGD. Moreover, by setting \(\lambda = \mathcal{O}(\frac{1}{\sqrt{T}})\), the term \(\frac{L}{\sqrt{T}K}\lambda^2 = \mathcal{O}(\frac{1}{KT^{3/2}})\) in the convergence rate, as generated by local SAM, can be neglected in comparison to the dominant term \(\mathcal{O}(\frac{1}{\sqrt{KT}})\). Furthermore, \method{} achieves a convergence rate of \(\mathcal{O}(\frac{1}{\sqrt{KT}})\), which has been demonstrated as the optimal rate in stochastic methods in DFL under general assumptions. Additionally, in the dominant term of the convergence rate \(\mathcal{O}(\frac{D}{\sqrt{KT}} + \frac{L}{\sqrt{T}K(1-\psi)^2}\sigma_l^2)\), it can be observed that better topological connectivity leads to a faster convergence rate. We will verify this conclusion in Section \ref{topoaware}. Moreover, a larger number of local epochs \(K\) also leads to a faster convergence rate, as confirmed in Section \ref{ablation_section} (see Figure \ref{fig:hyper} (b)).
\end{remark}

\begin{remark}\label{remark:2}
    Theorem \ref{the:opt_error} provides a general convergence bound for the \method{}. When \(\beta = 0\), it degrades to the vanilla DFedAvg method \cite{Sun2022Decentralized}. Similar to DFedAvg, it is influenced by the initialization bias \(D\) and the intrinsic variance \(\sigma_l\). However, with the adoption of initialization, the consistency term \(\Delta^T\) can assist in reducing the upper bound of the convergence rate. Furthermore, from Theorem \ref{the:opt_error}, it can be observed that a larger \(\beta\) leads to a smaller optimization error. On the other hand, \(\beta \leq \min\{\frac{\sqrt{10}(1-\psi)}{40}, \frac{\sqrt{5}}{30} \}\) provides an upper bound. By neglecting specific numerical relations and focusing on the associated relations, we can obtain an important conclusion: better-connected communication topology (implying a smaller \(1-\psi\)) corresponds to a smaller optimal value of \(\beta\) for the algorithm. Section \ref{topoaware} verifies our conclusion (see Figure \ref{fig:beta}).
\end{remark}

\begin{table*}[ht]  
\vspace{-0.3cm}
    \footnotesize  
    \centering 
    \caption{\small Top 1 test accuracy (\%) on two datasets in both IID and non-IID settings.}  
    \renewcommand{\arraystretch}{0.95}  
    \resizebox{0.9\textwidth}{!}{%
    \begin{tabular}{cccc!{\vrule width \lightrulewidth}ccc}  
        \toprule  
        \multirow{2}{*}{Algorithm}& \multicolumn{3}{c!{\vrule width \lightrulewidth}}{CIFAR-10}  & \multicolumn{3}{c}{CIFAR-100} \\  
        \cmidrule{2-7}  
        & \multicolumn{1}{c}{Dir 0.3} & \multicolumn{1}{c}{Dir 0.6} & \multicolumn{1}{c!{\vrule width \lightrulewidth}}{IID} &
         \multicolumn{1}{c}{Dir 0.3} & \multicolumn{1}{c}{Dir 0.6} & \multicolumn{1}{c}{IID} \\    
        \midrule  
        FedAvg      & 78.10 $\pm$  0.81 & 79.00 $\pm$  1.04 & 81.16 $\pm$  0.20
        & 54.50 $\pm$  0.70 & 55.69 $\pm$  0.72   & 56.94 $\pm$  0.41 \\ 

        FedSAM      & 80.22 $\pm$  0.70 & 81.43 $\pm$  0.28 & 82.83 $\pm$  0.23
        & 57.76 $\pm$  0.36 & 58.62 $\pm$  0.51   & 59.97 $\pm$  0.19 \\ 

        SCAFFOLD      & 78.39 $\pm$  0.35 & 79.85 $\pm$  0.18 & 81.75 $\pm$  0.23
        & 60.23 $\pm$  0.30 & 61.42 $\pm$  0.52   & 62.75 $\pm$  0.31 \\ 
        \midrule        
        D-PSGD       & 59.76 $\pm$  0.04 & 60.03 $\pm$  0.13 & 62.93 $\pm$  0.12
        & 55.68 $\pm$  0.20 & 56.68 $\pm$  0.10   & 57.60 $\pm$  0.12 \\    

        DFedAvg       & 77.25 $\pm$  0.12 & 77.83 $\pm$  0.11 & 79.97 $\pm$  0.08
        & 58.17 $\pm$  0.10 & 58.22 $\pm$  0.50   & 59.00 $\pm$  0.31 \\     
        
        DFedAvgM       & 79.30 $\pm$  0.24 & 80.66 $\pm$  0.07 & 82.72 $\pm$  0.20
        & 57.79 $\pm$  0.29 & 58.13 $\pm$  0.39   & 58.90 $\pm$  0.47 \\     
        
        DFedSAM       & 79.37 $\pm$  0.07 & 80.47 $\pm$  0.09 & 82.14 $\pm$  0.09
        &57.67 $\pm$  0.20  & 58.55 $\pm$  0.23   & 59.53 $\pm$  0.24 \\   
        \midrule
        \method{}-SGD      & 82.20 $\pm$ 0.20 & 82.72 $\pm$ 0.22 & 84.10 $\pm$ 0.24
        & 60.00 $\pm$ 0.15 & 60.65 $\pm$ 0.30 & 62.58 $\pm$ 0.21 \\

        \method{}-SAM      & \textbf{84.45} $\pm$ 0.19 & \textbf{84.70} $\pm$ 0.10 & \textbf{85.75} $\pm$ 0.20 
        & \textbf{61.22} $\pm$ 0.12 & \textbf{61.78} $\pm$ 0.22 & \textbf{63.55} $\pm$ 0.20 \\
        \midrule  
        \multirow{2}{*}{Algorithm}& \multicolumn{3}{c!{\vrule width \lightrulewidth}}{CIFAR-10}  & \multicolumn{3}{c}{CIFAR-100} \\  
        \cmidrule{2-7}  
        & \multicolumn{1}{c}{Path 2} & \multicolumn{1}{c}{Path 4} & \multicolumn{1}{c!{\vrule width \lightrulewidth}}{Path 6} &
         \multicolumn{1}{c}{Path 10} & \multicolumn{1}{c}{Path 20} & \multicolumn{1}{c}{Path 30} \\    
        \midrule  
        FedAvg & 69.01 $\pm$ 0.96 & 75.80 $\pm$ 1.01 & 76.76 $\pm$ 1.11 & 52.12 $\pm$ 0.70 & 55.70 $\pm$ 0.63 & 57.14 $\pm$ 0.26 \\

        FedSAM & 69.33 $\pm$ 1.32 & 75.78 $\pm$ 1.04 & 77.55 $\pm$ 0.68 & 52.46 $\pm$ 0.69 & 55.75 $\pm$ 0.55 & 57.60 $\pm$ 0.36 \\

        SCAFFOLD & 64.28 $\pm$ 2.12 & 78.08 $\pm$ 0.39 & 80.35 $\pm$ 0.19 & 54.17 $\pm$ 0.51 & 58.32 $\pm$ 0.67 & 60.50 $\pm$ 0.35 \\
          \midrule      
        D-PSGD & 59.53 $\pm$ 0.40 & 63.80 $\pm$ 0.22 & 64.86 $\pm$ 0.48 & 52.66 $\pm$ 0.58 & 55.79 $\pm$ 0.11 & 56.82 $\pm$ 0.35 \\

        DFedAvg & 74.73 $\pm$ 0.26 & 77.50 $\pm$ 0.25 & 79.00 $\pm$ 0.25 & 54.72 $\pm$ 0.19 & 57.57 $\pm$ 0.11 & 58.30 $\pm$ 0.24 \\
        
        DFedAvgM & 75.10 $\pm$ 0.27 & 79.50 $\pm$ 0.33 & 81.07 $\pm$ 0.27 & 48.19 $\pm$ 0.45 & 53.60 $\pm$ 0.64 & 53.95 $\pm$ 0.85 \\
        
        DFedSAM & 75.08 $\pm$ 0.11 & 79.85 $\pm$ 0.09 & 81.36 $\pm$ 0.11 & 53.86 $\pm$ 0.17 & 57.58 $\pm$ 0.22 & 58.80 $\pm$ 0.21 \\
        \midrule
        \method{}-SGD & 78.20 $\pm$ 0.24 & 81.54 $\pm$ 0.25 & 83.10 $\pm$ 0.30 & 55.03 $\pm$ 0.11 & 58.87 $\pm$ 0.13 & 59.75 $\pm$ 0.10 \\
        
        \method{}-SAM & \textbf{79.58} $\pm$ 0.18 & \textbf{83.45} $\pm$ 0.21 & \textbf{84.90} $\pm$ 0.18 & \textbf{56.68} $\pm$ 0.16 & \textbf{60.22} $\pm$ 0.24 & \textbf{61.57} $\pm$ 0.21 \\
        \bottomrule  
    \end{tabular}  }
    \vspace{-0.45cm}
    \label{ta:all_baselines}  
\end{table*}

\subsection{Generalization Analysis}

In this section, we primarily present a generalization analysis of our \method{} method under two gradient properties (bounded variance and Lipschitz continuity). We utilize uniform stability analysis, which is widely adopted in previous literature, to analyze the generalization error of \method{}. Next, we introduce the definition of uniform stability and then demonstrate the effectiveness of \method{}. All detailed proofs can be available in the Appendix \ref{appendix:generalization error}.

In DFL framework, we suppose there are $m$ clients participating in the training process as a set $\mathcal{C}=\{i\}_{i=1}^m$. Each client has a local dataset $\mathcal{S}_i=\{z_j\}_{j=1}^S$ with total $S$ data sampled from a specific unknown distribution $\mathcal{D}_i$. Now we define a re-sampled dataset $\widetilde{\mathcal{S}_i}$ which only differs from the dataset $\mathcal{S}_i$ on the $j^*$-th data. We replace the $\mathcal{S}_{i^*}$ with $\widetilde{\mathcal{S}}_{i^*}$ and keep other $m-1$ local dataset, which composes a new set $\widetilde{\mathcal{C}}$. $\mathcal{C}$ only differs from the $\widetilde{\mathcal{C}}$ at $j^*$-th data on the $i^*$-th client. Then, based on these two sets, \method{} could generate two solutions, $\bar{\mathbf{x}}^t$ and $\widetilde{\bar{\mathbf{x}}}^t$ respectively, after $t$ rounds. By bounding the difference according to these two models, we can obtain stability and generalization efficiency.

\begin{definition}
    (Uniform Stability \cite{Hardt2016train}) For these two models $\bar{\mathbf{x}}^T$ and $\widetilde{\bar{\mathbf{x}}}^T$ generated as introduced above, a general method satisfies $\epsilon$-uniformly stability if:
    \begin{equation}
        \operatorname*{sup}_{z_{j}\sim\{\mathcal{D}_{i}\}}\mathbb{E}[f(\bar{\mathbf{x}}^T;z_{j})-f(\widetilde{\bar{\mathbf{x}}}^T;z_{j})]\leq\epsilon.
    \end{equation}
    Moreover, if a general method satisfies $\epsilon$-uniformly stability, then its generalization error could also be bounded \cite{Zhang2021Stability, Hardt2016train} $$\mathcal{E}_G \leq \operatorname*{sup}_{z_{j}\sim\{\mathcal{D}_{i}\}}\mathbb{E}[f(\bar{\mathbf{x}}^T;z_{j})-f(\widetilde{\bar{\mathbf{x}}}^T;z_{j})]\leq\epsilon.$$
\end{definition}

\begin{theorem}\label{the:generation_err}
    Under Assumption \ref{as:bounded_stochastic_gradient}, \ref{as:bounded_heterogeneity}, and \ref{as:L_G-lip}, let all conditions in the optimization process be satisfied, let the learning rate be selected as $\eta = \mathcal{O}(\frac1t) = \frac{c}{t}$ where $c$ is a constant and $\beta \leq \frac{1-\psi}{4\sqrt{m} + 1-\psi}$, let $t_0$ be a specific round to firstly select the different data sample, and let $U = \sup_{\mathbf{x},z}\{ f ({\mathbf{x}}; z)\}$ be the upper bound, for arbitrary data sample $z$ followed the joint distribution $\mathcal{D}_i$, we have:
    \begin{equation*}
    \begin{aligned}
&\mathbb{E}\|f({\bar{\mathbf{x}}}^{T+1};z)-f(\widetilde{\bar{\mathbf{x}}}^{T+1};z)\| \leq  \frac{Ut_0}{S}+ \Big( \frac{2L_G(L_G+S\sigma_l)}{(1+2\beta)SL} \\
        &\quad +\frac{2L_G(1+\beta)K(\sigma_l + L_G)}{\alpha(1+2\beta)}C_\lambda \Big)\left(\frac{T}{t_{0}}\right)^{cKL}
    \end{aligned}
    \end{equation*}
    where \(S\) denotes the amount of data owned by each client, $\alpha = 1 - \frac{4\sqrt{m}\beta}{(1-\psi)(1-\beta)}, C_\lambda:=\ln\frac1\lambda\frac{\lambda^{\ln\frac1\lambda}}\lambda+\frac{\ln^2\frac1\lambda}{16\lambda}\lambda^{\frac{\ln\frac1\lambda}8}+\frac2{\lambda\ln\frac1\lambda}, (\lambda\neq 0); C_\lambda = 0 , (\lambda = 0).\\$ Furthermore, to minimize the stability errors, we can select the proper observation point $t_0 \!=\! T^{\frac{cKL}{1\!+\!cKL}} \!\!\left( (\frac{2L_G(L_G+S\sigma_l)}{(1\!+\!2\beta)SL} \!+\!\frac{2L_G(1+\beta)K(\sigma_l + L_G)}{\alpha(1+2\beta)}C_\lambda)\frac{ScKL}{U}\!\right)\!^{\frac{1}{1+cKL}}$ we then have
    \begin{equation*}
    \begin{aligned}
        &\mathbb{E}\|f({\bar{\mathbf{x}}}^{T+1};z)-f(\widetilde{\bar{\mathbf{x}}}^{T+1};z)\| 
        \leq 2T^{\frac{cKL}{1+cKL}}\Big( \frac{2(L_G+S\sigma_l)}{(1+2\beta)SL} \\
        &\quad +\frac{2(1+\beta)K(\sigma_l + L_G)}{\alpha(1+2\beta)}C_\lambda \Big)^{\frac{1}{1+cKL}}\left( \frac{U}{S}\right)^{\frac{cKL}{1+cKL}}
    \end{aligned}
    \end{equation*}
\end{theorem}



\begin{remark}\label{remark:gener_2}
    We provide further discussion on the impact of $\beta$ on the convergence rate and generalization performance of the algorithm. Firstly, from the negative coefficient in the last term of Theorem \ref{the:opt_error}, it can be inferred that a larger $\beta$ value will reduce the optimization error, thereby accelerating the convergence rate. As depicted in Figure \ref{fig:hyper} (a), with the increase in $\beta$, the convergence curve becomes steeper, demonstrating a faster convergence speed. Additionally, the expression of the generalization error derived from Theorem \ref{the:generation_err}, $( \frac{2(L_G+S\sigma_l)}{(1+2\beta)SL} +\frac{2(1+\beta)K(\sigma_l + L_G)}{\alpha(1+2\beta)}C_\lambda )^{\frac{1}{1+cKL}}$ indicates that as $\beta$ increases, the terms $\frac{1}{1+2\beta}$ and $\frac{1+\beta}{1+2\beta}$ decrease. This implies that with an increase in $\beta$, the generalization error of the algorithm will decrease, thereby enhancing its generalization performance. It is evident from Figure \ref{fig:hyper} (a) that as $\beta$ increases, the maximum value of the convergence curve (indicating generalization performance) also increases. In summary, a larger \( \beta \) will simultaneously enhance the convergence speed and generalization of the algorithm. Furthermore, from Theorem \ref{the:generation_err}, it can be observed that better topological connectivity (smaller \(C_\lambda\)) leads to a smaller generalization error, which is validated in Section \ref{topoaware} (see Figure \ref{fig:topo}).
\end{remark}

\begin{remark}
    Theorem \ref{the:generation_err} establishes the generalization error $\mathcal{E}_G$ of  \method{}. When \(\beta = 0\), it reduces to the vanilla DFedAvg \cite{Sun2022Decentralized}, the generalization error is given by \(\mathcal{O}\left((\frac{T}{S})^{\frac{cKL}{1+cKL}}(1+K C_\lambda)^{\frac{1}{1+cKL}}\right)\), which fills the gap in the generalization performance of DFedAvg \cite{Sun2022Decentralized}. From the generalization error, we mainly focus on the terms of the total number of data samples $S$, training length $T$ and $K$. Compared with the generalization bound of D-SGD proposed by \cite{Sun2022Stability}, \(\mathcal{O}((1+C_\lambda)T^{\frac{cL}{1+cL}})\), the local epochs \(K\) in DFedAvg can enhance its generalization, which is consistent with the observation in Figure \ref{fig:Compared_baselines-cifar10}\&\ref{fig:Compared_baselines-cifar100}. 
\end{remark}

%% file: tex/5.experiment.tex
\section{Experiment}
In this section, we conduct extensive experiments to verify the effectiveness of the proposed \method{}.

\begin{table*}[ht]
\vspace{-0.3cm}
\centering
\footnotesize
\caption{Communication rounds for each method achieving target accuracy on the CIFAR-10 dataset.}
\label{ta:CIFAR10-convergence}
\resizebox{0.95\textwidth}{!}{%
\begin{tabular}{cccc|ccc|ccc}
\toprule
\multirow{2}{*}{Methods} & \multicolumn{3}{c|}{Dir 0.3} & \multicolumn{3}{c|}{Dir 0.6} & \multicolumn{3}{c}{IID} \\  
\cmidrule{2-10}  
 & Acc@75      & Acc@77      & Acc@80      & Acc@75         & Acc@77         & Acc@80      & Acc@78         & Acc@80         & Acc@82    \\ \midrule  
FedAvg      & 141 (1.3$\times$)& 244 (1.7$\times$)& \textgreater 500 & 111 (1.4$\times$) & 166 (1.7$\times$)& 485 (1.0$\times$)& 150 (1.2$\times$)& 243 (2.0$\times$)& \textgreater 500 \\  
FedSAM    & 141 (1.3$\times$)& 202 (2.1$\times$) & 402 (1.2$\times$)& 121 (1.3$\times$) & 165 (1.7$\times$)& 296 (1.7$\times$)& 142 (1.2$\times$)& 199 (2.4$\times$) & 363 (0.8 $\times$)\\
SCAFFOLD    & 264 (0.7$\times$)& 356 (1.2$\times$) & \textgreater 500 & 202 (0.8$\times$) & 262 (1.1$\times$)& 470 (1.1$\times$)& 180 (1.0$\times$)& 273 (1.8$\times$) & \textgreater 500 \\
\midrule
D-PSGD      & \textgreater 500 & \textgreater 500 & \textgreater 500 & \textgreater 500 & \textgreater 500 & \textgreater 500 & \textgreater 500 & \textgreater 500 & \textgreater 500 \\  
DFedAvg     & 179 (1.0$\times$)& 419 (1.0$\times$)& \textgreater 500 & 152 (1.0$\times$) & 283 (1.0$\times$)& \textgreater 500 & 176 (1.0$\times$)& 479 (1.0$\times$)& \textgreater 500 \\  
DFedAvgM    & 93 (1.9$\times$)& 141 (3.0$\times$)& \textgreater 500 & 64 (2.4$\times$) & 99 (2.9$\times$)& 305 (1.6$\times$)& 59 (3.0$\times$)& 117 (4.1$\times$)& 303 (1.7$\times$)\\  
DFedSAM     & 187 (1.0$\times$)& 265 (1.6$\times$)& \textgreater 500 & 155 (1.0$\times$) & 203 (1.4$\times$)& 414 (1.2$\times$)& 143 (1.2$\times$)& 212 (2.3$\times$)& 452 (1.1$\times$)\\  
\midrule
\method{}-SGD    & \textbf{54 (3.3$\times$)}& 77 (5.4$\times$)& 146 (3.4$\times$)& \textbf{41 (3.7$\times$)}& 58 (4.9$\times$)& 110 (4.5$\times$)& \textbf{37 (4.8$\times$)}& \textbf{59 (8.1$\times$)}& 91 (5.5$\times$)\\  
\method{}-SAM & 57 (3.1$\times$) & \textbf{68 (6.2$\times$)}& \textbf{104 (4.8$\times$)}& 45 (2.9$\times$)& \textbf{53 (5.3$\times$)}& \textbf{82 (6.1$\times$)}& 48 (3.7$\times$)& 63 (7.6$\times$)& \textbf{90 (5.6$\times$)}\\  
\midrule
\multirow{2}{*}{Methods} & \multicolumn{3}{c|}{Path 2} & \multicolumn{3}{c|}{Path 4} & \multicolumn{3}{c}{Path 6} \\  
\cmidrule{2-10}  
 & Acc@55      & Acc@65      & Acc@70      & Acc@65         & Acc@70         & Acc@75      & Acc@70         & Acc@75         & Acc@80   \\ \midrule  
FedAvg      & 137 (0.5$\times$) & 283 (0.5$\times$)& \textgreater 500 & 113 (0.5$\times$)  & 180 (0.5$\times$) & 327 (0.6$\times$)& 129 (0.5$\times$) & 208 (0.5$\times$)& \textgreater 500 \\  
FedSAM    & 156 (0.4$\times$) & 313 (0.4$\times$)& 453 (0.5$\times$) & 119 (0.5$\times$)  & 172 (0.5$\times$) & 333 (0.6$\times$)& 132(0.4$\times$) & 225 (0.5$\times$)& \textgreater 500 \\ 
SCAFFOLD    & 183 (0.3$\times$)& 439 (0.3 $\times$) & \textgreater 500 & 122 (0.5$\times$) & 172 (0.5 $\times$)& 287 (0.7 $\times$)& 114 (0.5 $\times$)& 182 (0.6 $\times$) & 439 (1.1$\times$) \\
\midrule
D-PSGD      & 340 (0.2$\times$) & \textgreater 500 & \textgreater 500 & \textgreater 500 & \textgreater 500 & \textgreater 500 & \textgreater 500 & \textgreater 500 & \textgreater 500 \\  
DFedAvg     & 62 (1.0$\times$)& 139 (1.0$\times$) & 223 (1.0$\times$)& 59 (1.0$\times$) & 92 (1.0$\times$)& 187 (1.0$\times$)& 59 (1.0$\times$)& 113 (1.0$\times$) & \textgreater 500 \\  
DFedAvgM    & 126 (0.5$\times$)& 179 (0.8$\times$) & 244 (0.9$\times$)& 46 (1.3$\times$) & 64 (1.4$\times$)& 119 (1.6$\times$) & 32 (1.8$\times$)& 65 (1.7$\times$) & 257 (1.9$\times$)\\  
DFedSAM     & 75 (0.8$\times$)& 157 (0.9$\times$) & 248 (0.9$\times$) & 75 (0.8$\times$)& 111 (0.8$\times$)& 187 (1.0$\times$)& 84 (0.7$\times$)& 128 (0.9$\times$)& 328 (1.5$\times$)\\  
\midrule
\method{}-SGD    & \textbf{34 (1.8$\times$)} & \textbf{66 (2.1$\times$)} & \textbf{94 (2.4$\times$)} & \textbf{25 (2.4$\times$)} & \textbf{36 (2.6$\times$)} & \textbf{57 (3.3$\times$)} & \textbf{24 (2.5$\times$)} & \textbf{37 (3.1$\times$)}& 89 (5.6$\times$)\\  
\method{}-SAM & 44 (1.4$\times$) & 88 (1.6$\times$) & 117 (1.9$\times$) & 30 (2.0$\times$) & 41 (2.2$\times$) & 63 (3.0$\times$) & 31 (1.9$\times$) & 43 (2.6$\times$) & \textbf{83 (6.0$\times$)}\\  
\bottomrule
\end{tabular}}
\vspace{-0.3cm}
\end{table*}

\subsection{Experiment Setup}

\textbf{Dataset.} We evaluate the proposed \method{} on CIFAR-10\&100 datasets \cite{krizhevsky2009learning} in both IID and non-IID settings. To simulate non-IID data distribution among federated clients, we utilize the Dirichlet \cite{Hsu2019Measuring} and Pathological distribution \cite{Zhang2020Personalized}. Specifically, in the Dirichlet distribution, the local data of each client is partitioned by sampling label ratios from the Dirichlet distribution Dir($\alpha$). A smaller value of $\alpha$ indicates a higher degree of non-IID. In our experiments, we set $\alpha=0.3$ and $\alpha=0.6$ to represent different levels of non-IID. In the Pathological distribution, the local data of each client is partitioned by sampling label ratios from the Pathological distribution Path($\alpha$), where the value of $\alpha$ represents the number of classes owned by each client. For example, in the CIFAR-10 dataset, we set $\alpha = 2$ to indicate that each client possesses only 2 randomly selected classes out of the 10 available. In our experimental setup, we respectively set $\alpha=2$, $\alpha=4$, and $\alpha=6$ for the CIFAR-10 dataset, and $\alpha=10$, $\alpha=20$, and $\alpha=30$ for the CIFAR-100 dataset. 

\textbf{Baselines.} 
The baselines used for comparison include several state-of-the-art (SOTA) methods in both the CFL and DFL settings. Specifically, the centralized baselines include FedAvg \cite{mcmahan2017communication}, FedSAM \cite{caldarola2022improving,qu2022generalized}, and SCAFFOLD \cite{scaffold2020} . In the decentralized setting, D-PSGD \cite{lian2017can}, DFedAvg, and DFedAvgM \cite{Sun2022Decentralized}, along with DFedSAM \cite{shi2023improving}, are used for comparison. Note that both our baseline and proposed algorithms are designed to operate synchronously.

\textbf{Implementation Details.} The total number of clients is set to 100, with 10\% of the clients participating in the communication, creating a random bidirectional topology. For decentralized methods, all clients perform the local iteration step, while for centralized methods, only the participating clients perform the local update \cite{shi2023improving}. The local learning rate is initialized to 0.1 with a decay rate of 0.998 per communication round for all experiments. For SAM-based algorithms, such as DFedSAM and \method{}-SAM, we set the perturbation weight as $\lambda = 0.1$. As for $\beta$, we set $\beta=0.99$ for CIFAR-10 and $\beta=0.8$ for CIFAR-100 in the random topology. The maximum number of communication rounds is set to 500 for all experiments on CIFAR-10\&100. Additionally, all ablation studies are conducted on the CIFAR-10 dataset with a data partition method of Dir 0.3 and 500 communication rounds.

\textbf{Communication Configurations.} To ensure a fair comparison between decentralized and centralized approaches, we have implemented a dynamic and time-varying connection topology for DFL methods. This approach ensures that the number of connections in each round does not exceed the number of connections in the centralized server, thus enabling the matching of communication volume between the decentralized and centralized methods as \cite{dai2022dispfl}. 
To ensure a fair comparison, we regulate the number of neighbors for each client in DFL using the client participation rate. Here, we set each client to randomly select 10 neighbors during each communication round. Other communication topologies, such as Grid, are generated according to corresponding rules.

\subsection{Performance Evaluation}\label{section:exper-evaluation}

\begin{figure*}[h]
\begin{center}
\subfloat[CIFAR-10-Path]{
    	\includegraphics[width=1\textwidth]{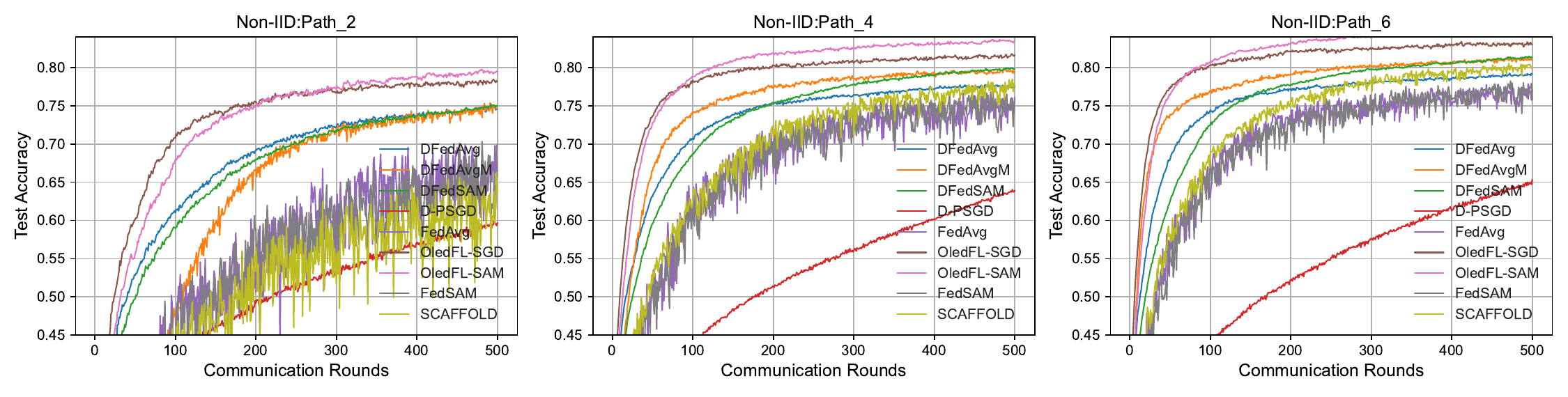}
    }
    \hfill
\subfloat[CIFAR-10-Dir]{
    	\includegraphics[width=1\textwidth]{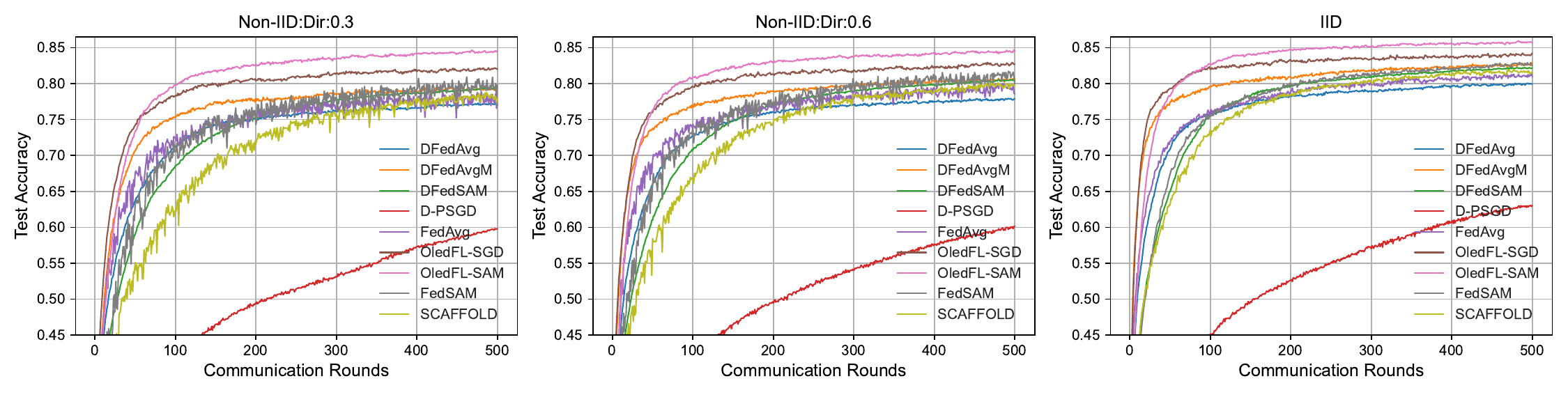}
    }
\end{center}
\caption{ \small Test accuracy of all baselines on CIFAR-10 in both IID and different non-IID settings.}
\label{fig:Compared_baselines-cifar10}
\end{figure*}

\begin{figure*}[ht]
\begin{center}
\subfloat[CIFAR-100-Path]{
    \includegraphics[width=1\textwidth]{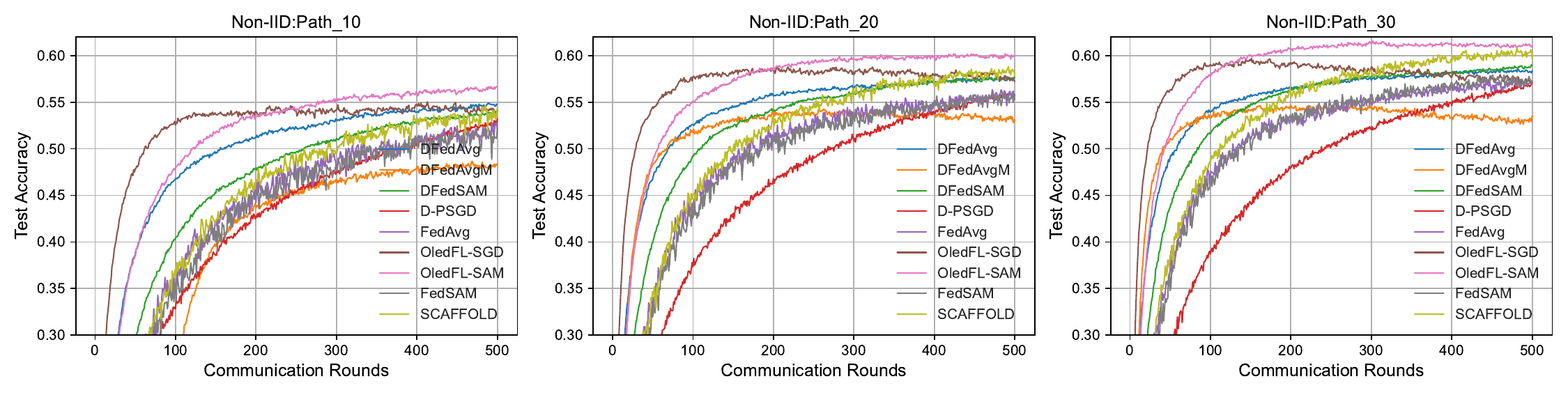}
}
    \hfill
\subfloat[CIFAR-100-Dir]{
    \includegraphics[width=1\textwidth]{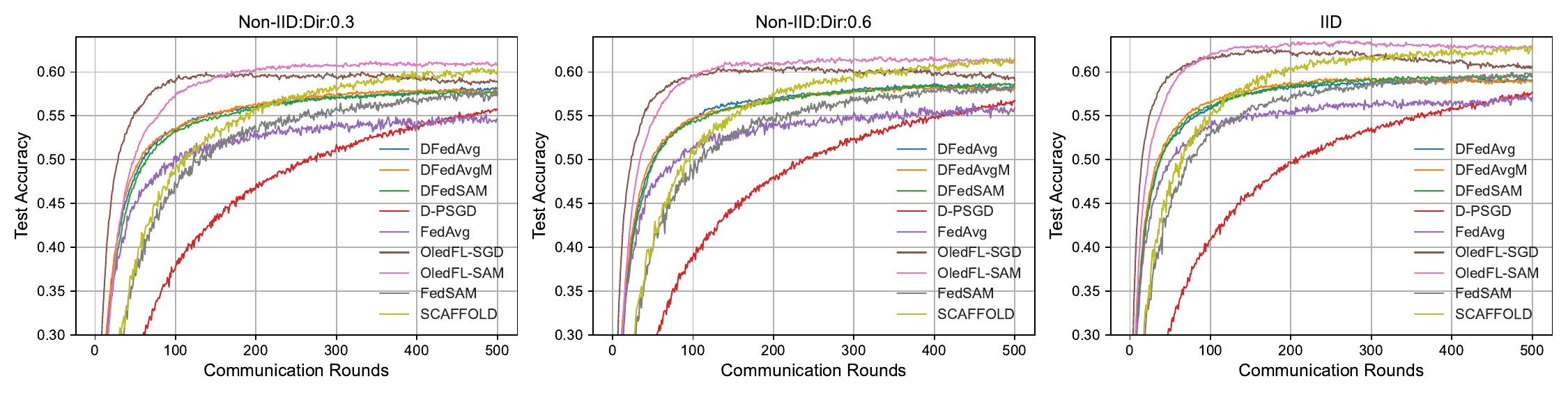}
}
\end{center}
\caption{ \small Test accuracy of all baselines on CIFAR-100 in both IID and different non-IID settings.}
\label{fig:Compared_baselines-cifar100}
\vspace{-0.3cm}
\end{figure*}

Figures \ref{fig:Compared_baselines-cifar10} and \ref{fig:Compared_baselines-cifar100} display the comparison between the baseline method and \method{} on the CIFAR10\&100 datasets under Dirichlet and Pathological distributions. From the figures, it is evident that \method{} can converge rapidly with fewer communication rounds, and \method{}-SAM also surpasses SCAFFOLD in terms of generalization performance. It is worth noting that, under the Pathological data distribution, CFL methods exhibit instability during training, whereas DFL methods provide a more stable training process.

\textbf{Performance Analysis.} In Table \ref{ta:all_baselines}, we conduct a series of experiments to compare the performance of our method with baseline methods. The results show that under various non-IID settings, our method consistently outperforms other methods, for example, on the CIFAR-10 dataset, our method outperforms the previous SOTA method DFedSAM by at least 4.5\% in both Dirichlet and Pathological distributions. Even in the IID case, our method also outperforms DFedSAM by at least 3.5\%. Similarly, these results are also observed on the more complex CIFAR-100 dataset, where compared to DFedSAM, our method provides at least a 3\% improvement under Dirichlet distribution and at least a 2.6\% improvement under Pathological distribution.

\textbf{Impact of non-IID levels (\(\alpha\)).} The robustness of \method{} in various non-IID settings is evident from Figure \ref{fig:Compared_baselines-cifar10}\& \ref{fig:Compared_baselines-cifar100} and Table \ref{ta:all_baselines}. A smaller value of alpha (\(\alpha\)) indicates a higher level of non-IID, which makes the task of optimizing a consensus problem more challenging. However, our algorithm consistently outperforms all baselines across different levels of non-IID. Furthermore, methods based on the SAM optimizer consistently achieve higher accuracy across different levels of non-IID. For instance, on the CIFAR-10 dataset, under the Dirichlet distribution, the performance of our algorithm \method{}-SAM only decreases by 1.3\% from the IID to Dir 0.3, which significantly outperforms DFedSAM (2.76\%), DFedAvgM (3.42\%), DFedAvg (2.72\%), D-PSGD (3.17\%), and FedAvg (3.06\%), respectively. This demonstrates the robustness of our method to heterogeneous data.


\begin{figure}[ht]
    \vspace{-0.2cm}
    \centering
    \subfloat{%
        \includegraphics[width=0.40\textwidth]{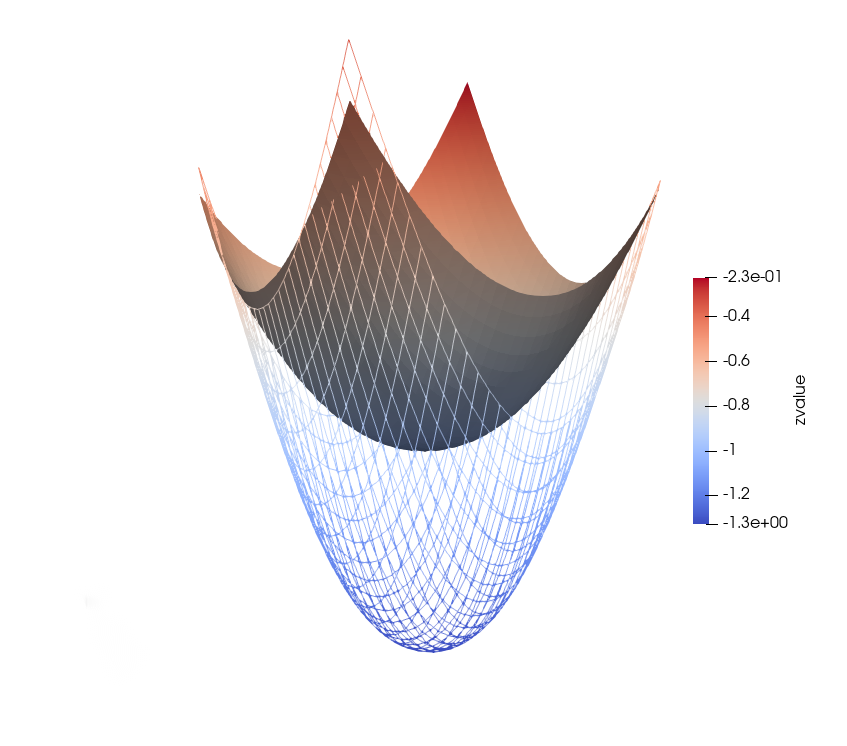}}
    \caption{ \small  The comparison of loss landscapes between DFedSAM and \method{}-SAM. Whereas the wireframe represents the loss landscape of \method{}-SAM, the surface represents the loss landscape of \method{}. It is clear that \method{}-SAM can find smoother minima.}
    \label{fig:comprision}
    \vspace{-0.3cm}
\end{figure}

\textbf{Impact of Ole Parameter $\beta$.} In Table \ref{ta:all_baselines} and Figure \ref{fig:Compared_baselines-cifar10}\&\ref{fig:Compared_baselines-cifar100}, we observe a significant improvement in generalization and convergence speed due to Ole initialization. Taking the example of CIFAR-10 under Dirichlet 0.3, parameter initialization helps the DFedAvg algorithm achieve an improvement of around 4\% in accuracy, and the DFedSAM algorithm achieves an improvement of around 5\% in accuracy. In terms of convergence speed, parameter initialization helps both the DFedAvg and DFedSAM algorithms achieve at least 3$\times$ speedup, and in some cases, even up to 8$\times$ speedup, which coincides with our theoretical analysis and demonstrates the efficacy of our proposed \method{} (see Remark \ref{remark:gener_2}).


\begin{figure*}[ht]
\begin{center}
\subfloat[DFedSAM-3D]{
    	\includegraphics[width=0.23\textwidth]{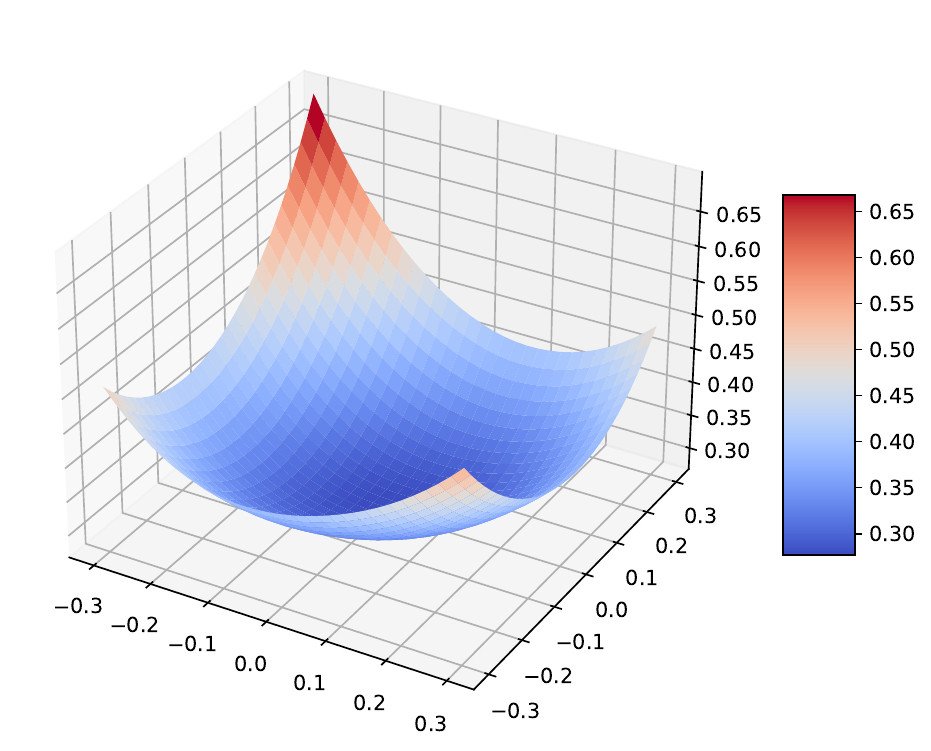}
    }
\subfloat[\method{}-SAM-3D]{
    \includegraphics[width=0.23\textwidth]{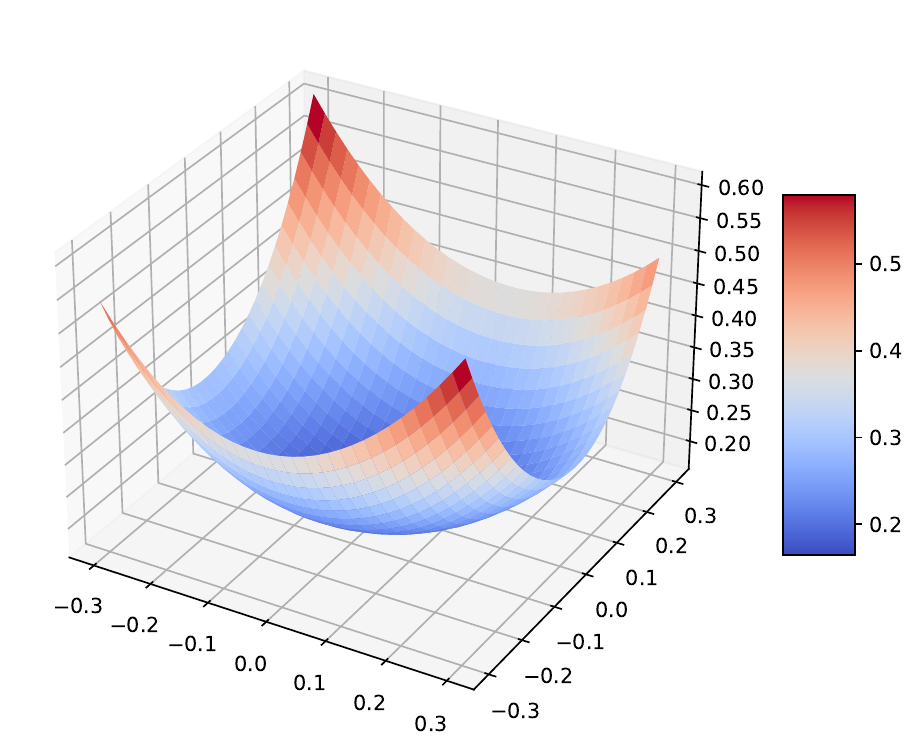}
}
\subfloat[DFedSAM-2D]{
    	\includegraphics[width=0.23\textwidth]{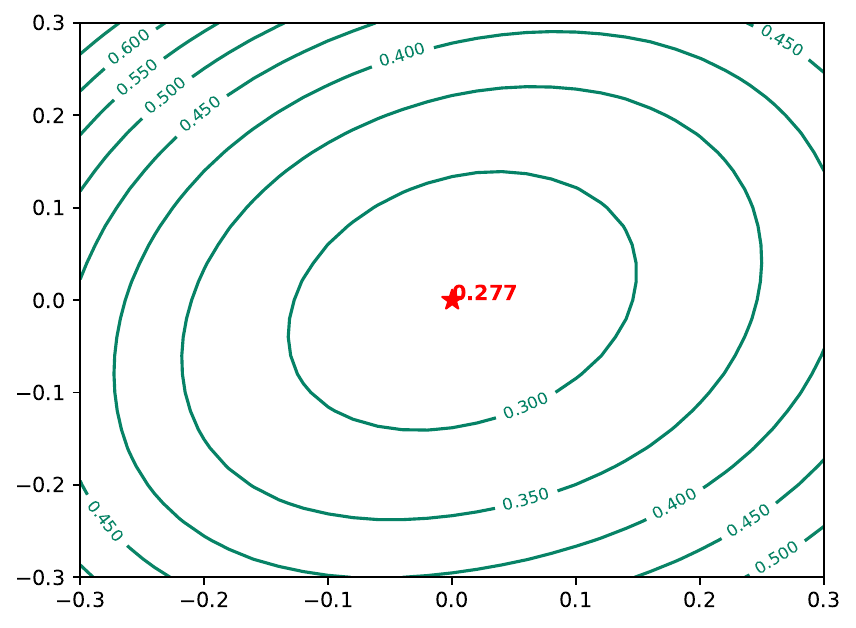}
    }
\subfloat[\method{}-SAM-2D]{
    \includegraphics[width=0.23\textwidth]{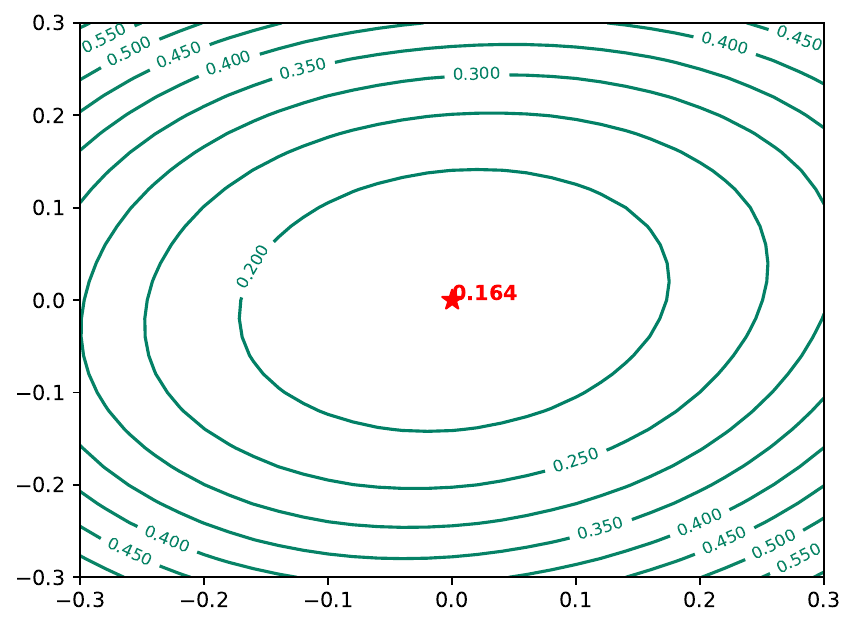}
}
\end{center}
\caption{ \small (a) and (b) depict the comparison of loss landscapes between DFedSAM and \method{}-SAM, while (c) and (d) show the contour plots of the loss landscapes of DFedSAM and \method{}-SAM. From (c) and (d), it can be observed that \method{}-SAM optimizes deeper than DFedSAM. As shown in Figure \ref{fig:comprision}, a comparison in (a) and (b) of Figure \ref{fig:losslandscape} indicates that \method{}-SAM is able to find a flatter loss surface.}
\label{fig:losslandscape}
\end{figure*}

\label{se:explain ole}\textbf{Explanation of the Effectiveness of Ole.} To explore the effectiveness of Ole, we plot the loss landscapes and corresponding contour maps of \method{}-SAM and DFedSAM on the CIFAR10 dataset under Dirichlet 0.3, as shown in Figure \ref{fig:losslandscape}. It is evident that \method{}-SAM can find parameters with lower loss values. Furthermore, from Figure \ref{fig:comprision}, it can be observed that the loss landscape of \method{}-SAM is flatter. According to the research findings of Keskar et al. \cite{Keskar2016Generalization} and Dziugaite et al. \cite{Dziugaite2017Generalization}, flatter loss surfaces lead to better generalization performance. This observation can explain why \method{}-SAM may exhibit superior generalization performance. This is also consistent with the theoretical analysis results in Remark \ref{remark:gener_2}.

\subsection{Topology-aware Performance}\label{topoaware}

Below, we explore the effects of topologies on different DFL methods on CIFAR-10 dataset with Dirichlet $\alpha=0.3$. 


\begin{table}[t]
\small
    \centering
    \caption{\small Top 1 test accuracy (\%) in various network topologies compared with decentralized algorithms on CIFAR-10.}
    \label{ta:topo}
    \renewcommand{\arraystretch}{1}
    \resizebox{0.45\textwidth}{!}{%
        \begin{tabular}{ccccc} 
            \toprule
            \multicolumn{1}{c}{Algorithm} & \multicolumn{1}{c}{Ring} & \multicolumn{1}{c}{Grid} & \multicolumn{1}{c}{Exp} & \multicolumn{1}{c}{Full}  \\ 
            \midrule
            D-PSGD        & 51.24 & 52.06 & 55.58 & 65.46            \\
            DFedAvg       & 63.30 & 74.72 & 77.11 & 78.52            \\
            DFedAvgM      & 65.49 & 76.89 & 78.01 & 80.14           \\
            DFedSAM       & 65.12 & 77.86 & 78.88 & 81.04            \\
            \midrule
            \method{}-SGD      & 71.78 & 81.05 & 78.73 & 81.52         \\
            \method{}-SAM  & \textbf{74.48} & \textbf{82.95} & \textbf{79.99} & \textbf{83.13}             \\
            \bottomrule
        \end{tabular} }
    \vspace{-0.5em}
\end{table}

\begin{figure}[ht]
\begin{center}
\subfloat{
    	\includegraphics[width=0.48\textwidth]{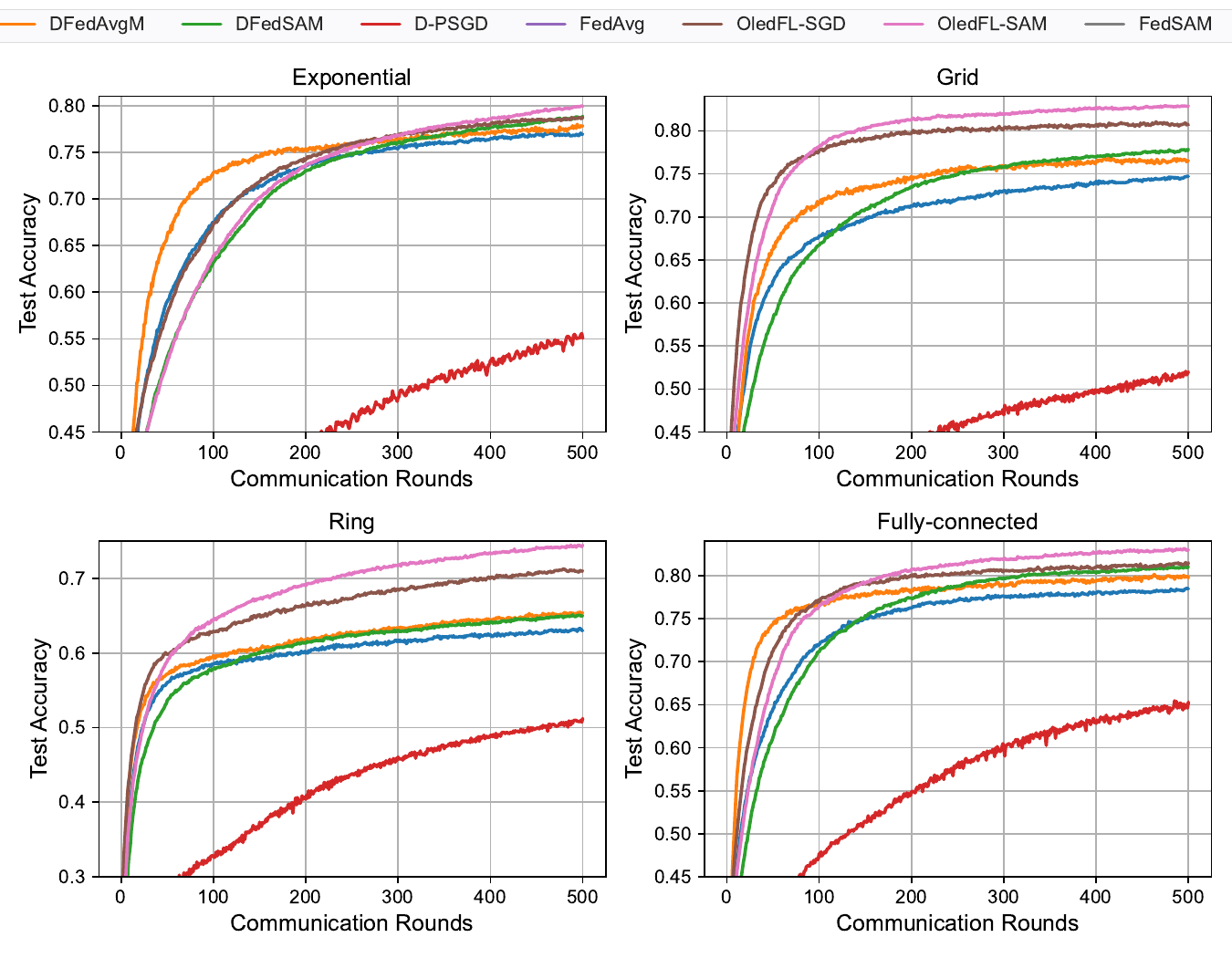}
    }
\end{center}
\caption{ \small Accuracy of different DFL algorithms with different decentralized topologies on the test dataset.}
\label{fig:topo}
 \vspace{-0.3cm}
\end{figure}

\begin{table}[ht]\small
    \centering
    \vspace{-1em}
    \caption{$\psi$ value of different topology. Here 0 means the extra term will disappear and N/A means the term will diverge.}
    \label{ta:topo_lambda}
    \renewcommand{\arraystretch}{1.2}
    \resizebox{0.4\textwidth}{!}{%
    \begin{tabular}{ccc} 
        \toprule
        \multicolumn{1}{c}{Topology} & \multicolumn{1}{c}{$\psi$} & \multicolumn{1}{c}{$\frac{1}{1-\psi}$}  \\ 
        \midrule
        Fully connected & 0 & 0 \\
        Exponential     & $1-\frac{2}{1+\ln{m}}$ & $\mathcal{O}(m)$ \\
        Grid            & $1-\frac{1}{m\ln{m}}$ & $\mathcal{O}(m\ln{m})$ \\
        Ring            & $1-\frac{16\pi^2}{3m^2}$ & $\mathcal{O}(m^2)$ \\
        \bottomrule
    \end{tabular} }
    \vspace{-0.5cm}
\end{table}

\textbf{Impact of Sparse Connectivity $\psi$.} Each client in the network only communicates with its predetermined neighbors, and the specific communication pattern is determined by the corresponding topology. In Table \ref{ta:topo_lambda}, the degree of sparse connectivity $\psi$ follows the order: Ring $>$ Grid $>$ Exponential $>$ Full-connected \cite{shi2023improving,zhu2022topology,Ying2021Exponential}. 
From Figure \ref{fig:topo} and Table \ref{ta:topo}, It can be observed that there is a general trend: as the sparse connectivity \(\psi\) decreases, the accuracy of our proposed algorithm on the test set increases. This can be attributed to the fact that a well-designed communication topology enables clients to obtain better initial optimization points through communication, leading to improved results. Furthermore, \method{} consistently achieves higher test set accuracy compared to other DFL baselines across various topologies. For example, Ole can enhance DFedSAM's performance by over 9\% on a Ring topology. Additionally, from Figure \ref{fig:topo}, it can be observed that the convergence rate and generalization performance of \method{}   generally increases with better topological connectivity, confirming the conclusion of Remarks \ref{remark:1}\&\ref{remark:gener_2}.


\begin{figure}[ht]
\begin{center}
\subfloat{
    	\includegraphics[width=0.30\textwidth]{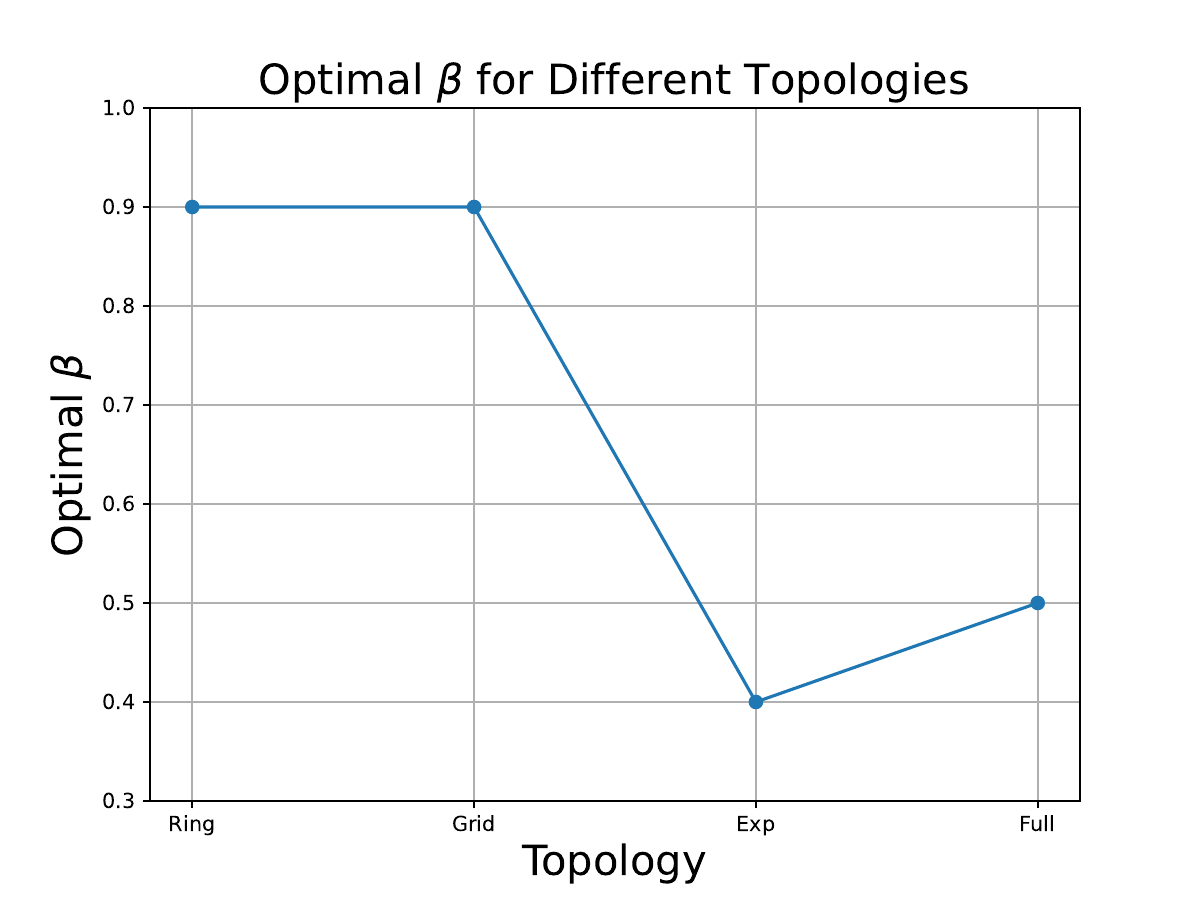}
    }
\end{center}
 \vspace{-1em}
\caption{ \small Optimal \(\beta\) for Different Topologies in \method{}. It can be observed that \method{}-SGD exhibits a decreasing relationship between the optimal $\beta$ under different topologies and the second largest eigenvalue $\psi$ of the communication topology. This confirms Remark \ref{remark:2} in Theorem \ref{the:opt_error}.}
\label{fig:beta}
 \vspace{-1em}
\end{figure}

\textbf{Relationship between \(\beta\) and $\psi$.} From Section \ref{section:Theoretical Analysis}, we have concluded that tighter topological connectivity (indicating smaller \(\psi\)) corresponds to smaller initial parameter values \(\beta\). To validate this theoretical result, we conduct experiments on the \method{} algorithm using the parameter set \(\beta = \{0.1, 0.2, \ldots, 0.9\}\) in each of the four topologies presented in Table \ref{ta:topo_lambda} to select the optimal \(\beta\) corresponding to the algorithm's performance. The results of the optimal \(\beta\) values under different topologies are presented in Figure \ref{fig:beta}. By comparing the \(\psi\)-\(\beta\) relationships in Figure \ref{fig:beta} with the corresponding values in Table \ref{ta:topo_lambda}, the experimental results validate our conclusion in Remark \ref{remark:2} of Theorem \ref{the:opt_error}.

\subsection{Ablation Study}\label{ablation_section}

We verify the influence of each component and hyperparameter in \method{}. All the ablation studies are conducted with the ``Random" topology, which is consistent with the communication configuration used in Section \ref{section:exper-evaluation}.

\begin{figure}[h]
\begin{center}
\subfloat{
    	\includegraphics[width=0.48\textwidth]{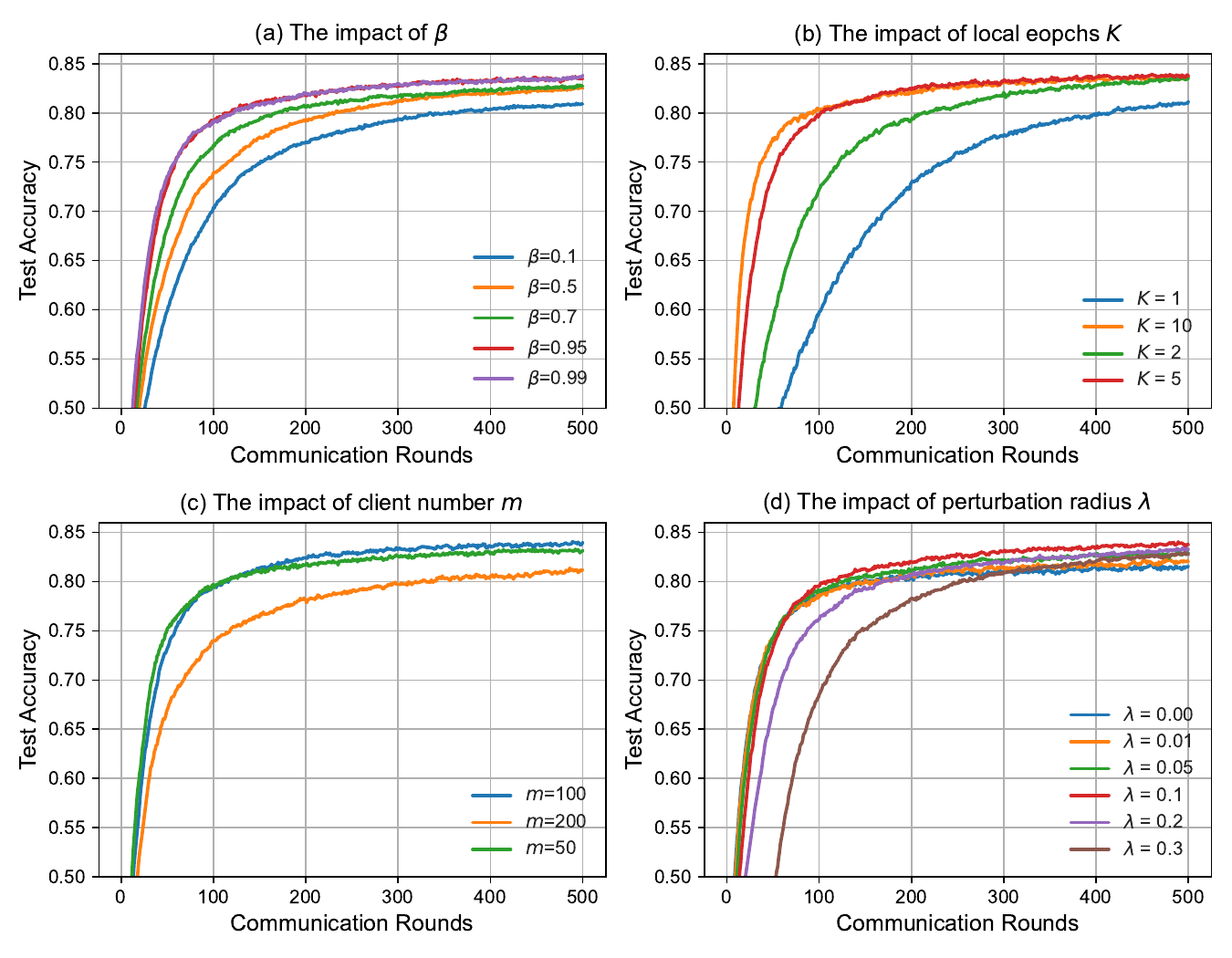}
    }
\end{center}
 \vspace{-0.4cm}
\caption{ Hyperparameter Sensitivity: local iterate $K$, Ole parameter $\beta$, number of participated clients $m$, perturbation radius $\lambda$.}
\label{fig:hyper}
 \vspace{-0.3cm}
\end{figure}

\textbf{Impact of Ole Parameter $\beta$.} The convergence curves under different Ole coefficients after 500 communication rounds are displayed in Figure \ref{fig:hyper} (a) on the CIFAR-10 dataset under the Dirichlet 0.3 distribution. A larger \(\beta\) value implies a greater deviation of each client's initialization point from the corresponding optimal point. We verify the performance curves of \method{}-SAM under the \(\beta\) parameter set \(\{0.1, 0.5, 0.7, 0.95, 0.99\}\). It can be seen that under the random bidirectional topology, the optimal \(\beta\) is 0.99. Furthermore, it is clear that as $\beta$ increases, the algorithm's convergence speed and generalization performance both improve. When $\beta > 1$, the algorithm diverges, which is consistent with the upper bound of $\beta$ obtained in the discussion of its relationship with CA in Section \ref{sec:oledfl alg}.

\textbf{Impact of Client Number $m$.} In Figure \ref{fig:hyper} (c), we present the performance with different numbers of participants, \(m = \{50, 100, 200\}\). We observe that, with an increase in the number of local data, \method{} achieves the best performance when \(m=50\) or 100, while a decrease in performance is observed when \(m\) is relatively large. This can be attributed to the fact that a smaller number of clients has a larger number of local training samples, leading to improved training performance.

\textbf{Impact of Perturbation Radius $\lambda$.} The perturbation radius \(\lambda\) is an additional factor influencing the convergence of \method{}. It controls the size of the perturbation radius, where a larger perturbation radius implies a more ambiguous direction of parameter descent, thereby affecting the convergence speed. We evaluate \method{}'s generalization performance using different values of \(\lambda\) from the set \(\{0, 0.01, 0.05, 0.1, 0.2, 0.3\}\). Figure \ref{fig:hyper} (d) illustrates the highest accuracy achieved when \(\lambda=0.1\). Additionally, we observe that a larger \(\lambda\) leads to a slower convergence speed, which is consistent with our previous analysis.

\textbf{Impact of local epochs $K$.} $K$ represents the number of optimization rounds performed by each client. A larger value of $K$ is more likely to lead to the "client drift" phenomenon, resulting in increased inconsistency between clients and thereby affecting the algorithm's performance. We assess the generalization performance of \method{} using different values of \(K\) from the set \(\{1, 2, 5, 10\}\). Figure \ref{fig:hyper} (b) demonstrates the highest accuracy achieved when \(K=5\). Additionally, we observe that a larger $K$ leads to faster convergence in Figure \ref{fig:hyper} (b), consistent with our theoretically proven $\mathcal{O}(\frac{1}{\sqrt{KT}})$.

%% file: tex/6.conclusion.tex
\section{Conlusion}

In this paper, we propose a plug-in method named \method{}, which integrates existing DFL optimizers to enhance the consistency among clients and significantly improve convergence speed and generalization performance at almost negligible computational cost. Theoretically, we are the first to conduct a joint analysis of algorithm convergence and generalization in the field of DFL, and we demonstrate the effectiveness of \method{} in reducing optimization error and generalization error. Moreover, a comprehensive explanation of the mechanism of action of Ole has been provided, encompassing intuitive, experimental, and theoretical perspectives. Furthermore, certain conclusions derived from theoretical deductions have been validated experimentally, thus offering additional insights into the Ole plugin methodology. Finally, extensive experiments on the CIFAR10\&100 datasets under Dirichlet and Pathological distributions demonstrate that \method{} can significantly reduce the performance gap between CFL and DFL, and even surpass CFL optimizers such as FedSAM and SCAFFOLD, which is crucial for further advancement in the field of DFL.

%% file: tex/8.proof.tex
\onecolumn
\appendix

In this part, we provide supplementary materials including the proof of the optimization and generalization error. Here’s the table of contents for the \textbf{Appendix}.

\begin{itemize}
    \item \textbf{Appendix} \ref{proof}: Proof of the theoretical analysis.
    \begin{itemize}
        \item \textbf{Appendix} \ref{appendix:optimization error}: Proof for optimization error.
        \item \textbf{Appendix} \ref{appendix:generalization error}: Proof for generalization error.
    \end{itemize}
\end{itemize}

\section{Supplementary Material for Experiment}\label{appendix:exper}

\section{Proof of the theoretical analysis.}\label{proof}

\subsection{Proofs for the Optimization Error}\label{appendix:optimization error}

In this part, we prove the training error for our proposed method. We assume the objective $ f({\bf x}):=\frac{1}{m}\sum_{i=1}^m f_i({\bf x}) $ is $L$-smooth w.r.t $\mathbf{x}$. Then we could upper bound the training error in the FL. Some useful notations in the proof are introduced in the Table \ref{ta:notation}. Then we introduce some important lemmas used in the proof.

\begin{table}[ht!]
\label{ta:notation}
  \caption{Some abbreviations of the used terms in the proof of bounded training error.}
  \centering  
  \begin{tabular}{ccc}
    \toprule  
    Notation & Formulation  & Description \\  
    \midrule  
    $\mathbf{x}_{i,k}^t$ & - & parameters at $k$-th iteration in round $t$ on client $i$ \\  
    $\mathbf{x}_{i}^t$ & - & global parameters in round $t$ on client $i$\\  
    $V_{1}^t$ & $\frac{1}{m}\sum_{i=1}^m\sum_{k=0}^{K-1}\mathbb{E}\| \mathbf{x}_{i,k}^t - \mathbf{x}_{i}^t \|^2$ & averaged norm of the local updates in round $t$ \\
    $V_{2}^t$ & $\mathbb{E}\| \bar{\mathbf{x}}^{t+1} - \bar{\mathbf{x}}^t \|^2 $ & norm of the average global updates in round $t$\\
    $\Delta^t$ & $ \frac1m\sum_{i=1}^m\mathbb{E}\|\mathbf{x}_{i,K}^{t-1}-\mathbf{x}_i^{t}\|^{2}$ & inconsistency / divergence term in round $t$  \\
    $D$ & $f(\mathbf{x}^{0})-f(\mathbf{x}^{\star})$  & bias between the initialization state and optimal \\
    \bottomrule  
  \end{tabular}  
\end{table}  

\subsubsection{Important Lemmas}
\begin{lemma}\label{lemma:1 bounded loacal updates}
    (Bounded local updates) We first bound the local training updates in the local training. Under the Assumptions stated, the averaged norm of the local updates of total $m$ clients could be bounded as:
    \begin{equation}
        \begin{aligned}
        V_1^t\leq 6K\beta^2\Delta^t+3K^2\eta^2\left(\sigma_l^2 + 6\lambda^2 + 6KG^2\right)+18K^3\eta^2B^2
        \frac{1}{m}\sum_{i=1}^m\mathbb{E}\|\nabla f(\mathbf{x}_i^t)\|^2.
        \end{aligned}
    \end{equation}
    \begin{proof}
        $V_{1}^t$ measures the norm of the local offset during the local training stage. It could be bounded by two major steps. Firstly, we bound the separated term on the single client $i$ at iteration $k$ as:
        \begin{equation}
            \begin{aligned}
                &\mathbb{E}_t\|\mathbf{x}_i^t-\mathbf{x}_{i,k}^t\|^2 \\
                &= \mathbb{E}_{t}\|\mathbf{x}_i^{t}-\mathbf{x}_{i,k-1}^{t}+\eta\left(g_{i,k-1}^{t}-\nabla f_{i}(\Breve{\mathbf{x}}_{i,k-1}^{t}) + \nabla f_{i}(\Breve{\mathbf{x}}_{i,k-1}^{t}) - \nabla f_{i}(\mathbf{x}_{i,k-1}^{t}) +\nabla f_{i}(\mathbf{x}_{i,k-1}^{t})-\nabla f_{i}(\mathbf{x}_i^{t})+\nabla f_{i}(\mathbf{x}_i^{t})\right)\|^{2}\\
                &\leq \left(1+\frac{1}{2K-1}\right)\mathbb{E}_t\| \mathbf{x}_i^{t}-\mathbf{x}_{i,k-1}^{t} \|^2 + 6\eta^2K\mathbb{E}_t\| \nabla f_{i}(\mathbf{x}_{i,k-1}^{t})-\nabla f_{i}(\mathbf{x}_i^{t})\|^2 + 6\lambda^2\eta^2\\
                & \quad + 6\eta^2K\mathbb{E}_t\| \nabla f_{i}(\mathbf{x}_i^{t})\|^2 +\eta^2\mathbb{E}_t\| g_{i,k-1}^{t}-\nabla f_{i}(\mathbf{x}_{i,k-1}^{t}) \|^2\\
                &\leq \left(1+\frac{1}{2K-1} + 6\eta^2KL^2\right)\mathbb{E}_t\| \mathbf{x}_i^{t}-\mathbf{x}_{i,k-1}^{t} \|^2 + 6\eta^2K\mathbb{E}_t\| \nabla f_{i}(\mathbf{x}_i^{t})\|^2 +\eta^2\sigma_l^2 + 6\lambda^2\eta^2 \\
                &\leq \left(1+\frac{1}{K-1}\right)\mathbb{E}_t\| \mathbf{x}_i^{t}-\mathbf{x}_{i,k-1}^{t} \|^2 + 6\eta^2K\mathbb{E}_t\| \nabla f_{i}(\mathbf{x}_i^{t})\|^2 +\eta^2\sigma_l^2 + 6\lambda^2\eta^2
            \end{aligned}
        \end{equation}
        where the learning rate is required $\eta\leq\frac{\sqrt{3}}{6KL}$.
        
        Computing the average of the separated term on client $i$, we have:
        \begin{equation}
            \begin{aligned}
                &\frac{1}{m}\sum_{i=1}^m\mathbb{E}_t\|\mathbf{x}_i^t-\mathbf{x}_{i,k}^t\|^2 \\
                &\leq \left(1+\frac{1}{K-1}\right)\frac{1}{m}\sum_{i=1}^m\mathbb{E}_t\| \mathbf{x}_i^{t}-\mathbf{x}_{i,k-1}^{t} \|^2 + 6\eta^2K \frac{1}{m}\sum_{i=1}^m \mathbb{E}_t\| \nabla f_{i}(\mathbf{x}_i^{t})\|^2 +\eta^2\sigma_l^2 + 6\lambda^2\eta^2\\
                &\leq \left(1+\frac{1}{K-1}\right)\frac{1}{m}\sum_{i=1}^m\mathbb{E}_t\| \mathbf{x}_i^{t}-\mathbf{x}_{i,k-1}^{t} \|^2 +\eta^2\sigma_l^2 + 6K\eta^2G^2+6K\eta^2B^2\frac{1}{m}\sum_{i=1}^m\mathbb{E}_t\|\nabla f(\mathbf{x}_i^t)\|^2 + 6\lambda^2\eta^2
            \end{aligned}
        \end{equation}
        
        Unrolling the aggregated term on iteration $k < K $. When local interval $K\geq 2, \left(1+\frac1{k-1}\right)^k \leq \left(1+\frac1{K-1}\right)^K \leq 4$. Then we have:
        \begin{equation}
            \begin{aligned}
                &\frac{1}{m}\sum_{i=1}^m\mathbb{E}_t\|\mathbf{x}_i^t-\mathbf{x}_{i,k}^t\|^2 \\
                &\leq \sum_{\tau = 0}^{k-1}\left(1+\frac{1}{K-1}\right)^{\tau}(\eta^2\sigma_l^2+ 6\lambda^2\eta^2 + 6K\eta^2G^2 + 6K\eta^2B^2\frac{1}{m}\sum_{i=1}^m\mathbb{E}_t\|\nabla f(\mathbf{x}_i^t)\|^2)\\
                &\quad + \left(1+\frac{1}{K-1}\right)^{k}\frac{1}{m}\sum_{i=1}^m\mathbb{E}_t\|\mathbf{x}_i^t-\mathbf{x}_{i,0}^t\|^2 \\
                &\leq 3(K-1)(\eta^2\sigma_l^2 + 6\lambda^2\eta^2 + 6K\eta^2G^2+6K\eta^2B^2\frac{1}{m}\sum_{i=1}^m\mathbb{E}_t\|\nabla f(\mathbf{x}_i^t)\|^2) + 4\beta^2\frac{1}{m}\sum_{i=1}^m\mathbb{E}_t\|\mathbf{x}_i^t-\mathbf{x}_{i,K}^{t-1}\|^2 \\
                &\leq 3\eta^2K(\sigma_l^2 + 6\lambda^2 + 6KG^2) + 18K^2\eta^2B^2\frac{1}{m}\sum_{i=1}^m\mathbb{E}_t\|\nabla f(\mathbf{x}_i^t)\|^2 + 4\beta^2\Delta^t
            \end{aligned}
        \end{equation}

        Summing the iteration on $K=0,1,2,\cdots ,K-1$.
        \begin{equation}
            \begin{aligned}
                \frac{1}{m}\sum_{i=1}^m\sum_{k=0}^{K-1}\mathbb{E}_t\|\mathbf{x}_i^t-\mathbf{x}_{i,k}^t\|^2 \leq 3\eta^2K^2(\sigma_l^2 +6\lambda^2 + 6KG^2) + 18 K^3\eta^2B^2\frac{1}{m}\sum_{i=1}^m\mathbb{E}_t\|\nabla f(\mathbf{x}_i^t)\|^2 + 4K\beta^2\Delta^t
            \end{aligned}
        \end{equation}
        This completes the proof.
    \end{proof}
\end{lemma}

\begin{lemma}\label{lemma2:bounded global updates}
    (Bounded global updates) Under assumptions stated above,  the norm of the global update of clients could be bounded as:
    \begin{equation}
        \begin{aligned}
            V_2^t \leq K\eta^2\sigma_l^2 + 2\frac{\eta^2}{m^2}\mathbb{E}\| \sum_{i=1}^m \sum_{k=0}^{K-1}\nabla f_i(\mathbf{x}_{i,k}^t)\|^2 + 2\eta^2\lambda^2
        \end{aligned}
    \end{equation}
    \begin{proof}
    \begin{equation}
        \begin{aligned}
            \mathbb{E}\|\bar{\mathbf{x}}^{t+1} - \bar{\mathbf{x}}^{t}\|^2 
            &= \mathbb{E}\| \frac{1}{m}\sum_{i=1}^m (\mathbf{x}_i^{t+1} - \mathbf{x}_i^{t})\|^2\\
            &= \mathbb{E}\| \frac{1}{m}\sum_{i=1}^m (\sum_{j=1}^m w_{i,j}\mathbf{x}_{j,K}^{t} - \mathbf{x}_i^{t})\|^2
        \end{aligned}
    \end{equation}
    As $\mathbf{W}$ is a doubly stochastic matrix, we have 
    $\sum_{i=1}^m\sum_{j=1}^m w_{i,j}^t \mathbf{x}_{j,K}^{t}= \sum_{j=1}^m\sum_{i=1}^m w_{i,j}^t \mathbf{x}_j^{t+\frac{1}{2}} = \sum_{j=1}^m\mathbf{x}_{j,K}^{t} = \sum_{i=1}^m\mathbf{x}_{i,K}^{t}$. Then we have
    \begin{equation}
        \begin{aligned}
            &\mathbb{E}\|\bar{\mathbf{x}}^{t+1} - \bar{\mathbf{x}}^{t}\|^2 \\
            &= \mathbb{E}\| \frac{1}{m}\sum_{i=1}^m (\mathbf{x}_{i,K}^{t} - \mathbf{x}_i^{t})\|^2\\
            &= \mathbb{E}\| \frac{1}{m}\sum_{i=1}^m \left(\sum_{k=0}^{K-1}\eta\mathbf{g}_{i,k}^t + \beta(\mathbf{x}_{i}^t - \mathbf{x}_{i,K}^{t-1}) \right)\|^2\\
            &= \mathbb{E}\| \frac{1}{m}\sum_{i=1}^m \left(\sum_{k=0}^{K-1}\eta(\mathbf{g}_{i,k}^t \pm \nabla f_i(\Breve{\mathbf{x}}_{i,k}^t) \pm \nabla f_i(\mathbf{x}_{i,k}^t)) + \beta(\mathbf{x}_{i}^t - \mathbf{x}_{i,K}^{t-1}) \right)\|^2\\ 
            &= \eta^2 \frac{1}{m}\sum_{i=1}^m \sum_{k=0}^{K-1}\mathbb{E}\|\mathbf{g}_{i,k}^t - \nabla f_i({\mathbf{x}}_{i,k}^t)\|^2 + 2\mathbb{E}\| \frac{1}{m}\sum_{i=1}^m \left(\eta\sum_{k=0}^{K-1}\nabla f_i(\mathbf{x}_{i,k}^t) + \beta(\mathbf{x}_{i}^t - \mathbf{x}_{i,K}^{t-1}) \right)\|^2 + 2\eta^2\lambda^2\\ 
            &\leq K\eta^2\sigma_l^2 + 2\mathbb{E}\| \frac{1}{m}\sum_{i=1}^m \left(\eta\sum_{k=0}^{K-1}\nabla f_i(\mathbf{x}_{i,k}^t) + \beta(\mathbf{x}_{i}^t - \mathbf{x}_{i,K}^{t-1}) \right)\|^2 + 2\eta^2\lambda^2
        \end{aligned}
    \end{equation}
    Because of $\sum_{i=1}^{m}\mathbf{x}_{i,K}^{t-1} = \sum_{i=1}^{m}\sum_{j=1}^m w_{i,j}\mathbf{x}_{j,K}^{t-1} = \sum_{i=1}^m\mathbf{x}_i^{t}$, we have $\sum_{i=1}^m(\mathbf{x}_{i}^t - \mathbf{x}_{i,K}^{t-1}) = 0$. Then, we get
    \begin{equation}
        \begin{aligned}
            \mathbb{E}\|\bar{\mathbf{x}}^{t+1} - \bar{\mathbf{x}}^{t}\|^2 
            &\leq K\eta^2\sigma_l^2 + 2\mathbb{E}\| \frac{1}{m}\sum_{i=1}^m \eta\sum_{k=0}^{K-1}\nabla f_i(\mathbf{x}_{i,k}^t) \|^2 + 2\eta^2\lambda^2\\
            &= K\eta^2\sigma_l^2 + 2\frac{\eta^2}{m^2}\mathbb{E}\| \sum_{i=1}^m \sum_{k=0}^{K-1}\nabla f_i(\mathbf{x}_{i,k}^t)\|^2 + 2\eta^2\lambda^2
        \end{aligned}
    \end{equation}
    Since $V_2^t = \mathbb{E}\|\bar{\mathbf{x}}^{t+1} - \bar{\mathbf{x}}^{t}\|^2$, we have completed the proof.
    \end{proof}
\end{lemma}

\begin{lemma}\label{mi}[Lemma 4, \cite{lian2017can}]
		For any $t\in \mathbb{Z}^+$, the mixing matrix ${\bf W}\in\ \mathbb{R}^m$ satisfies
		$\|{\bf W}^t-{\bf P}\|_{\emph{op}}\leq \psi^t,$
		where $\psi:=\max\{|\psi_2(\bf W)|,|\psi_m(\bf W)|\}$ and for a matrix ${\bf A}$, we denote its spectral norm as $\|{\bf A}\|_{\emph{op}}$. Furthermore, ${\bf 1}:=[1, 1, \ldots, 1 ]^{\top}\in \mathbb{R}^m$ and
		\begin{equation*}
			{\bf P}:=\frac{\mathbf{1}\mathbf{1}^{\top}}{m}\in \mathbb{R}^{m\times m}.
		\end{equation*}
\end{lemma}

In [Proposition 1, \cite{nedic2009distributed}], the author also proved that $\|{\bf W}^t-{\bf P}\|_{\textrm{op}}\leq C\psi^t$ for some $C>0$ that depends on the matrix.

\begin{lemma}\label{lemma4}
Let $\{\mathbf{x}_i^{t}\}_{t \ge 0}$ be generated by our proposed Algorithm for all $i \in \{1,2,...,m\}$ and any learning rate $\frac{1}{6KL} > \eta_l > 0 $, we have following bound:
    \begin{equation}
        \frac{1}{m}\sum_{i=1}^{m}\mathbb{E} [\|\mathbf{x}_i^{t} - \overline{\mathbf{x}}^{t} \|^2 ] \leq \frac{C_1^t}{(1-\psi)^2}.
    \end{equation}
        Where $C_1^t = 6K\beta^2\Delta^t+3K^2\eta^2\left(\sigma_l^2 +6\lambda^2 + 6KG^2\right) + 18K^3\eta^2B^2\frac{1}{m}\sum_{i=1}^m\mathbb{E}\|\nabla f(\mathbf{x}_i^t)\|^2 $.
    \begin{proof}
        Following [Lemma D.5, \cite{shi2023improving}], 
        we denote ${\bf Z}^{t}:=\begin{bmatrix}
            {\bf z}_1^{t},  {\mathbf z}_2^{t},
            \ldots,
            {\bf z}_m^{t}
        \end{bmatrix}^{\top}\in\mathbb{R}^{m\times d}$.
        With these notation, we have
        \begin{align}\label{xtglobal}
            {\bf X}^{t+1}={\bf W}{\bf Z}^{t}={\bf W}{\bf X}^{t}-{\bf \zeta}^t,
        \end{align}
        where ${\bf \zeta}^t:={\bf W}{\bf X}^{t}-{\bf W}{\bf Z}^{t}$.
        The iteration equation (\ref{xtglobal}) can be rewritten as the following expression
        \begin{align}\label{xtglobal2}
            {\bf X}^{t}={\bf W}^{t}{\bf X}^{0}-\sum_{j=0}^{t-1}{\bf W}^{t-1-j}{\bf \zeta}^j.
        \end{align}
        Obviously, it follows
        \begin{equation}\label{trans}
            {\bf W} {\bf P}= {\bf P} {\bf W}={\bf P}.
        \end{equation}
        According to Lemma \ref{mi}, it holds
        $$\|{\bf W}^t-{\bf P}\|\leq \psi^t.$$ Multiplying both sides of equation (\ref{xtglobal2}) with  ${\bf P}$ and using equation (\ref{trans}), we then get
        \begin{align}\label{xtglobal3}
            {\bf P}{\bf X}^{t}={\bf P}{\bf X}^{0}-\sum_{j=0}^{t-1}{\bf P}{\bf \zeta}^j=-\sum_{j=0}^{t-1}{\bf P}{\bf \zeta}^j,
        \end{align}
        where we used initialization ${\bf X}^{0}=\textbf{0}$.
        Then, we are led to
        \begin{equation}\label{xtglobal4}
            \begin{aligned}
                &\|{\bf X}^{t}-{\bf P}{\bf X}^{t}\|=\|\sum_{j=0}^{t-1}({\bf P}-{\bf W}^{t-1-j}){\bf \zeta}^j\|\\
                &\leq \sum_{j=0}^{t-1}\|{\bf P}-{\bf W}^{t-1-j}\|_{\textrm{op}}\|{\bf \zeta}^j\|\leq \sum_{j=0}^{t-1}\psi^{t-1-j}\|{\bf \zeta}^j\|.
            \end{aligned}
        \end{equation}
        With Cauchy inequality,
        \begin{align*}
            &\mathbb{E}\|{\bf X}^{t}-{\bf P}{\bf X}^{t}\|^2\leq \mathbb{E}(\sum_{j=0}^{t-1}\psi^{\frac{t-1-j}{2}}\cdot \psi^{\frac{t-1-j}{2}}\|{\bf \zeta}^j\|)^2\\
            &\leq (\sum_{j=0}^{t-1}\psi^{t-1-j})(\sum_{j=0}^{t-1} \psi^{t-1-j}\mathbb{E}\|{\bf \zeta}^j\|^2)
        \end{align*}
        Direct calculation gives us
        $$\mathbb{E}\|{\bf \zeta}^j\|^2\leq \|{\bf W}\|^2\cdot\mathbb{E}\|{\bf X}^{j}-{\bf Z}^{j}\|^2\leq \mathbb{E}\|{\bf X}^{j}-{\bf Z}^{j}\|^2.$$
        With Lemma \ref{lemma:1 bounded loacal updates}, for any $j$:
        \begin{equation}
            \begin{aligned}
                \mathbb{E}\|{\bf X}^{j}-{\bf Z}^{j}\|^2 \leq m(6K\beta^2\Delta^t+3K^2\eta^2\left(\sigma_l^2 +6\lambda^2 + 6KG^2\right) + 18K^3\eta^2B^2\frac{1}{m}\sum_{i=1}^m\mathbb{E}\|\nabla f(\mathbf{x}_i^t)\|^2)
            \end{aligned}
        \end{equation}
        Thus, we get:
        \begin{align*}
            \mathbb{E}\|{\bf X}^{t}-{\bf P}{\bf X}^{t}\|^2
            \leq \frac{mC_1^t}{(1-\psi)^2}.
        \end{align*}
        where $ C_1^t = 6K\beta^2\Delta^t+3K^2\eta^2\left(\sigma_l^2 +6\lambda^2 + 6KG^2\right) + 18K^3\eta^2B^2\frac{1}{m}\sum_{i=1}^m\mathbb{E}\|\nabla f(\mathbf{x}_i^t)\|^2 $.
        
        The fact that ${\bf X}^{t}-{\bf P}{\bf X}^{t}=\left(
        \begin{array}{c}
            {\bf x}_1^t- \overline{{\bf x}^t}\\
            {\bf x}_1^t- \overline{{\bf x}^t} \\
            \vdots \\
            {\bf x}_m^t- \overline{{\bf x}^t} \\
        \end{array}
        \right)
        $ then proves the result.
    \end{proof}	
\end{lemma}

\begin{lemma}\label{lemma:bouned divergence term}
    (Bounded divergence term) Under assumptions stated above, $\beta^2 \leq min\{\frac{(1-\psi)^2}{30K} ,\frac{1}{60(2 +\frac{K}{(1-\psi)^2})}\}$ and $\eta\leq \frac{1}{KL}$, The divergence term $\Delta^t$ could be bounded as the recursion of:
    \begin{equation*}
        \Delta^t\leq \frac{\Delta^t - \Delta^{t+1}}{1-\gamma} + \frac{5}{\mu(1-\gamma)}C_2^t
    \end{equation*}
    Where $\gamma\mu = 5\beta^2\left( 25 + \frac{6K}{(1-\psi)^2}\right), \mu = 1-\frac{30K}{(1-\psi)^2}\beta^2, C_2^t = 3\left(\frac{2K}{(1-\psi)^2} + 1\right)\eta^2K\sigma_l^2  + 4\left(\frac{9K}{(1-\psi)^2} + 1\right)\eta^2K^2G^2 + 2\frac{\eta^2}{m^2}\mathbb{E}\|\sum_{i=1}^m\sum_{k=0}^{K-1}\nabla f_i(\mathbf{x}_{i,k}^t) \|^2 + 4\left(\frac{9K}{(1-\psi)^2} + 1\right)\eta^2K^2B^2\frac{1}{m}\sum_{i=1}^m\mathbb{E}\|\nabla f(\mathbf{x}_i^t)\|^2 + 4\left( 2 + \frac{9K^2}{(1-\psi)^2} \right)\lambda^2 \eta^2.$
    \begin{proof}
        The divergence term measures the inconsistency level in the FL framework. According to the local updates, we have the following recursive formula:
        \begin{equation}\label{18}
            \begin{aligned}
                \underbrace{\mathbf{x}_i^{t+1}-\mathbf{x}_{i,K}^{t}}_{\textit{local bias in round t}+1}=\beta\underbrace{(\mathbf{x}_{i,K}^{t-1}-\mathbf{x}_i^{t})}_{\textit{local bias in nound t}}+(\mathbf{x}_i^{t+1}-\mathbf{x}_i^{t})+\sum_{k=0}^{K-1}\eta \mathbf{g}_{i,k}^{t}.
            \end{aligned}
        \end{equation}
        By taking the squared norm and expectation on both sides, we have:
        \begin{equation*}
            \begin{aligned}
                &\mathbb{E}\| \mathbf{x}_i^{t+1}-\mathbf{x}_{i,K}^{t} \|^2\\
                &=  \mathbb{E}\|\beta(\mathbf{x}_{i,K}^{t-1}-\mathbf{x}_i^{t}) + \mathbf{x}_i^{t+1} - \bar{\mathbf{x}}^{t+1} + \bar{\mathbf{x}}^{t} - \mathbf{x}_i^{t} + \bar{\mathbf{x}}^{t+1} - \bar{\mathbf{x}}^{t} + \sum_{k=0}^{K-1}\eta \mathbf{g}_{i,k}^{t}\|^2 \\
                &\leq 5\beta^2\mathbb{E}\|\mathbf{x}_{i,K}^{t-1}-\mathbf{x}_i^{t}\|^2  + 5\underbrace{\mathbb{E}\| \bar{\mathbf{x}}^{t+1} - \bar{\mathbf{x}}^{t} \|^2}_{V_2^t} + 5\mathbb{E}\| \mathbf{x}_i^{t+1} - \bar{\mathbf{x}}^{t+1} \|^2 + 5\mathbb{E}\|  \bar{\mathbf{x}}^{t} - \mathbf{x}_i^{t} \|^2 + 5\mathbb{E}\| \sum_{k=0}^{K-1}\eta \mathbf{g}_{i,k}^{t} \|^2 
            \end{aligned}
        \end{equation*}
        The second term in the above inequality is $V_2^t$ we have bounded in lemma \ref{lemma2:bounded global updates}. The third and fourth terms have been bounded in Lemma \ref{lemma4}. Then we bound the stochastic gradients term. We have:
        \begin{equation*}
            \begin{aligned}
                \mathbb{E}\| \sum_{k=0}^{K-1}\eta \mathbf{g}_{i,k}^{t} \|^2 &= \eta^2 \mathbb{E}\| \sum_{k=0}^{K-1} \mathbf{g}_{i,k}^{t} \|^2\\
                &\leq \eta^2\mathbb{E}\| \sum_{k=0}^{K-1}\left( \mathbf{g}_{i,k}^{t} -\nabla f_i(\Breve{\mathbf{x}}_{i,k}^{t})\right)\|^2 + 2\eta^2\mathbb{E}\| \sum_{k=0}^{K-1}\nabla f_i(\mathbf{x}_{i,k}^{t})\|^2 + 2\eta^2\lambda^2\\
                &\leq \eta^2K\sigma_l^2 + 2\eta^2 K \sum_{k=0}^{K-1}\mathbb{E}\| \nabla f_i(\mathbf{x}_{i,k}^{t}) - \nabla f_i(\mathbf{x}_{i}^{t}) + \nabla f_i(\mathbf{x}_{i}^{t})\|^2  + 2\eta^2\lambda^2\\
                &\leq \eta^2K\sigma_l^2 + 4\eta^2 KL^2 \sum_{k=0}^{K-1}\mathbb{E}\| \mathbf{x}_{i,k}^{t} - \mathbf{x}_{i}^{t}\|^2  + 4\eta^2K\sum_{k=0}^{K-1}\mathbb{E}\|\nabla f_i(\mathbf{x}_{i}^{t})\|^2 + 2\eta^2\lambda^2
            \end{aligned}
        \end{equation*}
        Taking the average on client i, we have:
        \begin{equation*}
            \begin{aligned}
                \frac{1}{m}\sum_{i=1}^m \mathbb{E}\| \sum_{k=0}^{K-1}\eta \mathbf{g}_{i,k}^{t} \|^2 
                &\leq \eta^2K\sigma_l^2 + 4\eta^2 KL^2\underbrace{\frac{1}{m}\sum_{i=1}^{m} \sum_{k=0}^{K-1}\mathbb{E}\| \mathbf{x}_{i,k}^{t} - \mathbf{x}_{i}^{t}\|^2}_{V_1^t}  + \frac{4\eta^2K}{m}\sum_{i=1}^m\sum_{k=0}^{K-1}\mathbb{E}\|\nabla f_i(\mathbf{x}_{i}^{t})\|^2  + 2\eta^2\lambda^2\\
                &\leq \eta^2K\sigma_l^2 + 4\eta^2 KL^2{V_1^t}  + 4\eta^2K^2G^2 + 4\eta^2K^2B^2\frac{1}{m}\sum_{i=1}^m\mathbb{E}\|\nabla f(\mathbf{x}_{i}^{t})\|^2  + 2\eta^2\lambda^2
            \end{aligned}
        \end{equation*}
        Combining this and the squared norm inequality, we have:
        \begin{equation*}
            \begin{aligned}
                &\Delta^{t+1} = \frac{1}{m}\sum_{i=1}^m\mathbb{E}\| \mathbf{x}_i^{t+1}-\mathbf{x}_{i,K}^{t}\|^2\\
                &\leq 5\beta^2\Delta^t  + 5{V_2^t} + \frac{5C_1^t}{(1-\psi)^2} + \frac{5C_1^{t+1}}{(1-\psi)^2} + 5\frac{1}{m}\sum_{i=1}^m \mathbb{E}\| \sum_{k=0}^{K-1}\eta \mathbf{g}_{i,k}^{t} \|^2 \\
                &\leq 5\beta^2\Delta^t + 5\left(\frac{1}{(1-\psi)^2} + 4\eta^2KL^2\right)V_1^t + 5\frac{1}{(1-\psi)^2}V_1^{t+1} +20\eta^2\lambda^2\\
                &\quad + 10\eta^2K\sigma_l^2 + 10\frac{\eta^2}{m^2}\mathbb{E}\|\sum_{i=1}^m\sum_{k=0}^{K-1}\nabla f_i(\mathbf{x}_{i,k}^t)\|^2 + 20\eta^2K^2G^2 + 20\eta^2K^2B^2\frac{1}{m}\sum_{i=1}^m\mathbb{E}\|\nabla f(\mathbf{x}_i^t)\|^2\\
                &\overset{(a)}\leq 5\left( 1 + 24\eta^2K^2L^2 + \frac{6K}{(1-\psi)^2}\right)\beta^2\Delta^t + \frac{30K}{(1-\psi)^2}\beta^2\Delta^{t+1} + 15\left(\frac{2K}{(1-\psi)^2} + 1\right)\eta^2K\sigma_l^2 + 20\left( \frac{9K^2}{(1-\psi)^2} + 2\right)\lambda^2\eta^2 \\
                &\quad + 20\left(\frac{9K}{(1-\psi)^2} + 1\right)\eta^2K^2G^2 + 10\frac{\eta^2}{m^2}\mathbb{E}\|\sum_{i=1}^m\sum_{k=0}^{K-1}\nabla f_i(\mathbf{x}_{i,k}^t) \|^2 + 20\left(\frac{9K}{(1-\psi)^2} + 1\right)\eta^2K^2B^2\frac{1}{m}\sum_{i=1}^m\mathbb{E}\|\nabla f(\mathbf{x}_i^t)\|^2
            \end{aligned}
        \end{equation*}
        Where (a) uses Lemma \ref{lemma:1 bounded loacal updates} and $\eta$ is a very small value, i.e., we have omitted terms of $\mathcal{O}(\eta^4)$. Let $C_2^t$ denote the term that is independent of $\Delta^t$,i.e., $C_2^t = 3\left(\frac{2K}{(1-\psi)^2} + 1\right)\eta^2K\sigma_l^2  + 4\left(\frac{9K}{(1-\psi)^2} + 1\right)\eta^2K^2G^2 + 2\frac{\eta^2}{m^2}\mathbb{E}\|\sum_{i=1}^m\sum_{k=0}^{K-1}\nabla f_i(\mathbf{x}_{i,k}^t) \|^2 + 4\left(\frac{9K}{(1-\psi)^2} + 1\right)\eta^2K^2B^2\frac{1}{m}\sum_{i=1}^m\mathbb{E}\|\nabla f(\mathbf{x}_i^t)\|^2 + 4\left( 2 + \frac{9K^2}{(1-\psi)^2} \right)\lambda^2 \eta^2$. Then we have:
        \begin{equation*}
            \begin{aligned}
                \left(1-\frac{30K}{(1-\psi)^2}\beta^2\right)\Delta^{t+1} \leq 5\left( 1 + 24\eta^2K^2L^2 + \frac{6K}{(1-\psi)^2}\right)\beta^2\Delta^t + 5C_2^t
            \end{aligned}
        \end{equation*}
        Let $\frac{30K}{(1-\psi)^2}\beta^2 < 1$ where $\beta < \frac{1-\psi}{\sqrt{30K}} $ . Let $\mu = \left(1-\frac{30K}{(1-\psi)^2}\beta^2\right)$ and dividing both sides by $\mu$, we obtain:
        \begin{equation*}
            \Delta^{t+1} \leq \frac{5\left( 1 + 24\eta^2K^2L^2 + \frac{6K}{(1-\psi)^2}\right)\beta^2}{\left(1-\frac{30K}{(1-\psi)^2}\beta^2\right)}\Delta^t + \frac{5}{\left(1-\frac{30K}{(1-\psi)^2}\beta^2\right)}C_2^t
        \end{equation*}
        Utilizing the learning rate condition $\eta \leq \frac{1}{KL}$, we obtain $\eta^2K^2L^2\leq1$, then let $ \gamma = \frac{5\beta^2\left( 25 + \frac{6K}{(1-\psi)^2}\right)}{1-\frac{30K}{(1-\psi)^2}\beta^2} <1$ where $\beta^2 \leq \frac{1}{60(2+\frac{K}{(1-\psi)^2})}$, thus we add $(1-\gamma)\Delta^t$ on both sides and get the recursive formulation:
        \begin{equation*}
            \Delta^t\leq \frac{\Delta^t - \Delta^{t+1}}{1-\gamma} + \frac{5}{\mu(1-\gamma)}C_2^t
        \end{equation*}
    \end{proof}
\end{lemma}

\subsubsection{Expanding the Smoothness Inequality for the Non-convex Objective}
For the non-convex and $L$-smooth function $f$ , we firstly expand the smoothness inequality at round $t$ as:
\begin{equation}\label{eq:main}
    \begin{aligned}
        &\mathbb{E}[f(\bar{\mathbf{x}}^{t+1})-f(\bar{\mathbf{x}}^t)] \\
        &\leq \mathbb{E}\langle\nabla f(\bar{\mathbf{x}}^t),\bar{\mathbf{x}}^{t+1}-\bar{\mathbf{x}}^t\rangle+\frac L2\underbrace{\mathbb{E}\|\bar{\mathbf{x}}^{t+1}-\bar{\mathbf{x}}^t\|^2}_{V_2^t}\\
        & = \mathbb{E}\langle\nabla f(\bar{\mathbf{x}}^t),\frac{1}{m}\sum_{i=1}^m (\mathbf{x}_{i,K}^{t} - \mathbf{x}_i^{t})\rangle+\frac{LV_2^t}{2}\\
        & = \mathbb{E}\langle\nabla f(\bar{\mathbf{x}}^t),\frac{1}{m}\sum_{i=1}^m \left(\mathbf{x}_{i,K}^{t} - \mathbf{x}_{i,0}^{t} + \beta(\mathbf{x}_{i}^{t} - \mathbf{x}_{i,K}^{t-1})\right)\rangle+\frac{LV_2^t}{2}\\
        & = -\eta\mathbb{E}\langle\nabla f(\bar{\mathbf{x}}^t),\frac{1}{m}\sum_{i=1}^m \sum_{k=0}^{K-1}\nabla f_i(\mathbf{x}_{i,k}^t) -\frac{1}{m}\sum_{i=1}^m \sum_{k=0}^{K-1}\nabla f_i(\bar{\mathbf{x}}^t) + K\nabla f(\bar{\mathbf{x}}^t)\rangle+\frac{LV_2^t}{2}\\
        &= -\eta K \mathbb{E}\| \nabla f(\bar{\mathbf{x}}^t) \|^2  + \mathbb{E}\langle\sqrt{\eta K}\nabla f(\bar{\mathbf{x}}^t),\sqrt{\frac{\eta}{K}}\frac{1}{m}\sum_{i=1}^m \sum_{k=0}^{K-1}\left(\nabla f_i(\bar{\mathbf{x}}^t)-\nabla f_i(\mathbf{x}_{i,k}^t)\right)\rangle+\frac{LV_2^t}{2}\\
        & \leq -\eta K \mathbb{E}\| \nabla f(\bar{\mathbf{x}}^t) \|^2 + \frac{\eta K }{2}\mathbb{E}\| \nabla f(\bar{\mathbf{x}}^t) \|^2 + \frac{\eta}{2m}\sum_{i=1}^m \sum_{k=0}^{K-1}\mathbb{E}\|\nabla f_i(\bar{\mathbf{x}}^t)-\nabla f_i(\mathbf{x}_{i,k}^t) \|^2 \\
        &\quad -\frac{\eta}{2Km^2}\mathbb{E}\| \sum_{i=1}^m \sum_{k=0}^{K-1}\nabla f_i(\mathbf{x}_{i,k}^t)\|^2 +\frac{LV_2^t}{2}\\
        & \leq - \frac{\eta K }{2}\mathbb{E}\| \nabla f(\bar{\mathbf{x}}^t) \|^2 + \frac{\eta L^2}{2}\underbrace{\frac{1}{m}\sum_{i=1}^m \sum_{k=0}^{K-1}\mathbb{E}\|\bar{\mathbf{x}}^t-\mathbf{x}_{i,k}^t \|^2}_{V_1^t}  -\frac{\eta}{2Km^2}\mathbb{E}\| \sum_{i=1}^m \sum_{k=0}^{K-1}\nabla f_i(\mathbf{x}_{i,k}^t)\|^2 +\frac{LV_2^t}{2}\\
        & \leq - \frac{\eta K }{2}\mathbb{E}\| \nabla f(\bar{\mathbf{x}}^t) \|^2 + \frac{\eta L^2V_1^t}{2}  -\frac{\eta}{2Km^2}\mathbb{E}\| \sum_{i=1}^m \sum_{k=0}^{K-1}\nabla f_i(\mathbf{x}_{i,k}^t)\|^2 +\frac{LV_2^t}{2}\\
    \end{aligned}
\end{equation}

According to Lemma \ref{lemma:1 bounded loacal updates} and lemma \ref{lemma2:bounded global updates} to bound the $V_1^t$ and $V_2^t$ , we can get the following recursive formula:
\begin{equation*}
    \begin{aligned}
        &\mathbb{E}[f(\bar{\mathbf{x}}^{t+1})-f(\bar{\mathbf{x}}^t)] \\
        &\leq - \frac{\eta K }{2}\mathbb{E}\| \nabla f(\bar{\mathbf{x}}^t) \|^2 + \frac{\eta L^2}{2}\left( 6K\beta^2\Delta^t+3K^2\eta^2\left(\sigma_l^2+6KG^2 +6\lambda^2\right)+18K^3\eta^2B^2
        \frac{1}{m}\sum_{i=1}^m\mathbb{E}\|\nabla f(\mathbf{x}_i^t)\|^2 \right)  \\
        &\quad + \frac{\eta^2 L K}{2}\sigma_l^2 +\left(\frac{\eta^2 L}{m^2}-\frac{\eta}{2Km^2}\right)\mathbb{E}\|\sum_{i=1}^m \sum_{k=0}^{K-1}\nabla f_i(\mathbf{x}_{i,k}^t) \|^2 + L\eta^2\lambda^2\\
        &\leq - \frac{\eta K }{2}\mathbb{E}\| \nabla f(\bar{\mathbf{x}}^t) \|^2 + 3\eta L^2K\beta^2\Delta^t  + 9\eta^3 K^3 L^2G^2 + 9\eta^3K^3L^2B^2
        \frac{1}{m}\sum_{i=1}^m\mathbb{E}\|\nabla f(\mathbf{x}_i^t)\|^2   \\
        &\quad + \frac{\eta^2 L K}{2}\left(1+3\eta LK\right)\sigma_l^2 +\left(\frac{\eta^2 L}{m^2}-\frac{\eta}{2Km^2}\right)\mathbb{E}\|\sum_{i=1}^m \sum_{k=0}^{K-1}\nabla f_i(\mathbf{x}_{i,k}^t) \|^2 + (1 + 9K^2\eta L)L\eta^2\lambda^2\\
    \end{aligned}
\end{equation*}
Furthermore, with Lemma \ref{lemma4}, we can get:
\begin{equation*}
    \begin{aligned}
        \frac1m\sum_{i=1}^m\mathbb{E}\left\|\nabla f(\mathbf{x}_i^t)\right\|^2 &\leq 2L^2\frac{\sum_{i=1}^m\left\|\mathbf{x}_i^t-\overline{\mathbf{x}}^t\right\|^2}m + 2\mathbb{E}\left\|\nabla f(\overline{\mathbf{x}}^t)\right\|^2 \\
        &\leq 2L^2\frac{C_1^t}{(1-\psi)^2} + 2\mathbb{E}\left\|\nabla f(\overline{\mathbf{x}}^t)\right\|^2 
    \end{aligned}
\end{equation*}
Where $C_1^t = 6K\beta^2\Delta^t+3K^2\eta^2\left(\sigma_l^2 +6\lambda^2 +6KG^2\right) + 18K^3\eta^2B^2\frac{1}{m}\sum_{i=1}^m\mathbb{E}\|\nabla f(\mathbf{x}_i^t)\|^2$, Therefore, we have:
\begin{equation*}
    \begin{aligned}
        \frac1m\sum_{i=1}^m\mathbb{E}\left\|\nabla f(\mathbf{x}_i^t)\right\|^2 \leq  \frac{8KL^2\beta^2\Delta^t + 6\eta^2K^2L^2(\sigma_l^2 + 4KG^2) + 2(1-\psi)^2\mathbb{E}\left\|\nabla f(\overline{\mathbf{x}}^t)\right\|^2}{(1-\psi)^2 - 24\eta^2K^3B^2L^2}
    \end{aligned}
\end{equation*}
And then, (\ref{eq:main}) can be represented as
\begin{equation}\label{main:2}
    \begin{aligned}
        &\mathbb{E}[f(\bar{\mathbf{x}}^{t+1})-f(\bar{\mathbf{x}}^t)]\\ 
        &\leq - \eta K(\frac{1 }{2} - 18\eta^2K^2L^2B^2)\mathbb{E}\| \nabla f(\bar{\mathbf{x}}^t) \|^2 + \eta L^2K\beta^2\left( 3 + \frac{72\eta^2K^3L^2B^2}{(1-\psi)^2}\right)\Delta^t  + 9\eta^3 K^3 L^2\left( 1 + \frac{24\eta^2K^3L^2B^2}{(1-\psi)^2} \right)G^2 \\
        &\quad + \eta^2LK\left( \frac{1}{2} + \frac{3}{2}\eta LK + \frac{54\eta^3K^4L^3B^2}{(1-\psi)^2}\right)\sigma_l^2 +\left(\frac{\eta^2 L}{m^2}-\frac{\eta}{2Km^2}\right)\mathbb{E}\|\sum_{i=1}^m \sum_{k=0}^{K-1}\nabla f_i(\mathbf{x}_{i,k}^t) \|^2  + (1 + 9K^2\eta L)L\eta^2\lambda^2\\
    \end{aligned}
\end{equation}
Where we use vary small $\eta$ \cite{shi2023improving}. Furthermore, in Lemma \ref{lemma:bouned divergence term}, $\mu -\mu\gamma = 1-\frac{60K\beta^2}{(1-\psi)^2} -75\beta^2$. By setting $\beta^2 \leq \min\{\frac{(1-\psi)^2}{240K}, \frac{1}{300}\}$, we can obtain $\mu -\mu\gamma > \frac{1}{2}$. Next, with Lemma \ref{lemma:bouned divergence term}, we get:

\begin{equation}\label{omit}
    \begin{aligned}
        \eta L^2K\beta^2\left( 3 + \frac{72\eta^2K^3L^2B^2}{(1-\psi)^2}\right)\Delta^t & \leq  \eta L^2K\beta^2\left( 3 + \frac{72\eta^2K^3L^2B^2}{(1-\psi)^2}\right)\frac{\Delta^t - \Delta^{t+1}}{1-\gamma} + \mathcal{O}(\eta^3)
    \end{aligned}
\end{equation}

We omit the terms of $\mathcal{O}(\eta^3)$ in (\ref{omit}) and substitute (\ref{omit}) into (\ref{main:2}) to obtain:
\begin{equation*}
    \begin{aligned}
        &\mathbb{E}[f(\bar{\mathbf{x}}^{t+1})-f(\bar{\mathbf{x}}^t)]\\ 
        &\leq - \eta K(\frac{1 }{2} - 18\eta^2K^2L^2B^2)\mathbb{E}\| \nabla f(\bar{\mathbf{x}}^t) \|^2 + \eta L^2K\beta^2\left( 3 + \frac{72\eta^2K^3L^2B^2}{(1-\psi)^2}\right)\frac{\Delta^t - \Delta^{t+1}}{1-\gamma}  + 9\eta^3 K^3 L^2\left( 1 + \frac{24\eta^2K^3L^2B^2}{(1-\psi)^2} \right)G^2 \\
        &\quad + \eta^2LK\left( \frac{1}{2} + \frac{3}{2}\eta LK + \frac{54\eta^3K^4L^3B^2}{(1-\psi)^2}\right)\sigma_l^2 +\left(\frac{\eta^2 L}{m^2}-\frac{\eta}{2Km^2}\right)\mathbb{E}\|\sum_{i=1}^m \sum_{k=0}^{K-1}\nabla f_i(\mathbf{x}_{i,k}^t) \|^2   + (1 + 9K^2\eta L)L\eta^2\lambda^2
    \end{aligned}
\end{equation*}
Firstly, to remove the gradient term, we follow the \cite{Yang2021achieving, scaffold2020} and let $\frac{\eta^2 L}{m^2}-\frac{\eta}{2Km^2} \leq 0$, then learning rate $\eta \leq \frac{1}{2KL}$. Then, according to the \cite{Yang2021achieving}, there exists a positive constant $\kappa \in (0, 1)$ such that $\frac{1 }{2} - 18\eta^2K^2L^2B^2 \geq \kappa >0$ when $\eta \leq \frac{1}{6KLB}$. Also, when $\eta \leq \frac{1}{K^{3/2}LB}$, we have $\eta^2K^3L^2B^2 \leq 1$. Therefore, we have:
\begin{equation}\label{main:main}
    \begin{aligned}
        \kappa \eta K \mathbb{E}\| \nabla f(\bar{\mathbf{x}}^t) \|^2 &\leq  \mathbb{E}[f(\bar{\mathbf{x}}^t) - f(\bar{\mathbf{x}}^{t+1})] + \eta L^2 K\beta^2\left(3 + \frac{72}{(1-\psi)^2}\right)\frac{\Delta^t - \Delta^{t+1}}{1-\gamma} \\
        &\quad + 9\eta^3 K^3 L^2\left( 1 + \frac{24}{(1-\psi)^2} \right)G^2 + \eta^2LK\left( 2 + \frac{36}{(1-\psi)^2}\right)\sigma_l^2 + (1 + 9K^2\eta L)L\eta^2\lambda^2
    \end{aligned}
\end{equation}

\subsubsection{Proof of Theorem 1}

\begin{theorem}
Under Assumption \ref{as:smoothness} - \ref{as:bounded_heterogeneity}, let the learning rate satisfy $\eta \leq  \frac{1}{K^{3/2}LB}$ where $K \geq 2$, let the relaxation coefficient $\beta \leq min\{\frac{\sqrt{10}(1-\psi)}{40}, \frac{\sqrt{5}}{30} \}$ , and after training $T$ rounds, the averaged model parameters generated by our proposed algorithm satisfies: 
\begin{equation*}
    \begin{aligned}
        \frac{1}{T}\sum_{t=0}^{T-1}\mathbb{E}\| \nabla f(\bar{\mathbf{x}}^t) \|^2 
        &\leq  \frac{\mathbb{E}[f(\bar{\mathbf{x}}^0) - f(\bar{\mathbf{x}}^{T})]}{\kappa \eta K T} + 9\eta^2 K^2 L^2\left( 1 + \frac{24}{(1-\psi)^2} \right)\frac{G^2}{\kappa} + \eta L\left( 2 + \frac{36}{(1-\psi)^2}\right)\frac{\sigma_l^2}{\kappa}  \\
        &\quad+ (1 + 9K^2\eta L)L\eta^2\lambda^2 -  L^2 \beta^2\left(3 + \frac{72}{(1-\psi)^2}\right)\frac{\Delta^{T}}{\kappa(1-\gamma)T}
    \end{aligned}
\end{equation*}
Where $\kappa \in (0,1)$ is a constant. 

Further, by selecting the proper learning rate $\eta = \mathcal{O}(\frac{1}{\sqrt{KT}})$ and let $D = f(\bar{\mathbf{x}}^0) - f(\bar{\mathbf{x}}^*)$ as the initialization bias, thenthe averaged model parameters $\bar{\mathbf{x}}^t$ satisfies:
\begin{equation*}
    \frac{1}{T}\sum_{t=0}^{T-1}\mathbb{E}\| \nabla f(\bar{\mathbf{x}}^t) \|^2 = \mathcal{O}(\frac{D}{\sqrt{KT}} + \frac{KL^2}{T(1-\psi)^2}G^2 + \frac{L}{\sqrt{T}K(1-\psi)^2}\sigma_l^2 + \frac{L}{TK}\lambda^2)
\end{equation*}

\begin{proof}
    According to the expansion of the smoothness inequality (\ref{main:main}), we have:
    \begin{equation*}
        \begin{aligned}
        \kappa \eta K \mathbb{E}\| \nabla f(\bar{\mathbf{x}}^t) \|^2 &\leq  \mathbb{E}[f(\bar{\mathbf{x}}^t) - f(\bar{\mathbf{x}}^{t+1})] + \eta L^2 K\beta^2\left(3 + \frac{72}{(1-\psi)^2}\right)\frac{\Delta^t - \Delta^{t+1}}{1-\gamma} \\
        &\quad + 9\eta^3 K^3 L^2\left( 1 + \frac{24}{(1-\psi)^2} \right)G^2 + \eta^2LK\left( 2 + \frac{36}{(1-\psi)^2}\right)\sigma_l^2 + (1 + 9K^2\eta L)L\eta^2\lambda^2
        \end{aligned}
    \end{equation*}
    Taking the accumulation from $0$ to $T - 1$, we have:
    \begin{equation*}
        \begin{aligned}
            &\frac{1}{T}\sum_{t=0}^{T-1}\mathbb{E}\| \nabla f(\bar{\mathbf{x}}^t) \|^2\\
            &\leq \frac{\mathbb{E}[f(\bar{\mathbf{x}}^0) - f(\bar{\mathbf{x}}^{T})]}{\kappa \eta K T} + L^2 \beta^2\left(3 + \frac{72}{(1-\psi)^2}\right)\frac{\Delta^0 - \Delta^{T}}{\kappa(1-\gamma)T} \\
            &\quad + 9\eta^2 K^2 L^2\left( 1 + \frac{24}{(1-\psi)^2} \right)\frac{G^2}{\kappa} + \eta L\left( 2 + \frac{36}{(1-\psi)^2}\right)\frac{\sigma_l^2}{\kappa} + (1 + 9K^2\eta L)L\eta\frac{\lambda^2}{\kappa} \\
            &\leq \frac{\mathbb{E}[f(\bar{\mathbf{x}}^0) - f(\bar{\mathbf{x}}^{T})]}{\kappa \eta K T} + 9\eta^2 K^2 L^2\left( 1 + \frac{24}{(1-\psi)^2} \right)\frac{G^2}{\kappa} + \eta L\left( 2 + \frac{36}{(1-\psi)^2}\right)\frac{\sigma_l^2}{\kappa} + (1 + 9K^2\eta L)L\eta\frac{\lambda^2}{\kappa}\\
            &\quad -  L^2 \beta^2\left(3 + \frac{72}{(1-\psi)^2}\right)\frac{\Delta^{T}}{\kappa(1-\gamma)T}
        \end{aligned}
    \end{equation*}
    We select the learning rate $\eta = \mathcal{O}(\frac{1}{\sqrt{KT}})$ and let $D = f(\bar{\mathbf{x}}^0) - f(\bar{\mathbf{x}}^*)$ as the initialization bias, then we have:
    \begin{equation*}
        \frac{1}{T}\sum_{t=0}^{T-1}\mathbb{E}\| \nabla f(\bar{\mathbf{x}}^t) \|^2 = \mathcal{O}(\frac{D}{\sqrt{KT}} + \frac{KL^2}{T(1-\psi)^2}G^2 + \frac{L}{\sqrt{T}K(1-\psi)^2}\sigma_l^2 + \frac{L}{\sqrt{T}K}\lambda^2)
    \end{equation*}
    This completes the proof.
\end{proof}
\end{theorem}

\subsection{Proofs for the Generalization Error}\label{appendix:generalization error}

In this part, we prove the generalization error for our proposed method. We assume the objective function $f$ is $L$-smooth and $L_G$-Lipschitz as defined in \cite{Hardt2016train,Zhou2021towards}. We follow the uniform stability to upper bound the generalization error in the DFL. 

We suppose there are $m$ clients participating in the training process as a set $\mathcal{C}=\{i\}_{i=1}^m$. Each client has a local dataset $\mathcal{S}_i=\{z_j\}_{j=1}^S$ with total $S$ data sampled from a specific unknown distribution $\mathcal{D}_i$. Now we define a re-sampled dataset $\widetilde{\mathcal{S}_i}$ which only differs from the dataset $\mathcal{S}_i$ on the $j^*$-th data. We replace the $\mathcal{S}_{i^*}$ with $\widetilde{\mathcal{S}}_{i^*}$ and keep other $m-1$ local dataset, which composes a new set $\widetilde{\mathcal{C}}$. $\mathcal{C}$ only differs from the $\widetilde{\mathcal{C}}$ at $j^*$-th data on the $i^*$-th client. Then, based on these two sets, our method could generate two output models, $\bar{\mathbf{x}}^t$ and $\widetilde{\bar{\mathbf{x}}}^t$ respectively, after $t$ training rounds. We first introduce some notations used in the proof of the generalization error.

\begin{table}[h]
\label{ta:notation_gener}
  \caption{Some abbreviations of the used terms in the proof of bounded stability error.}
  \centering  
  \begin{tabular}{ccc}
    \toprule  
    Notation & Formulation  & Description \\  
    \midrule  
    $\bar{\mathbf{x}}^t$ & - & parameters trained with set $\mathcal{C}$ \\  
    $\widetilde{\bar{\mathbf{x}}}^t$ & - & parameters trained with set $\widetilde{\mathcal{C}}$\\  
    $\delta_k^t$ & $ \frac1m\sum_{i=1}^m\mathbb{E}\|\mathbf{x}_{i,k}^{t}-\widetilde{\mathbf{x}}_{i,k}^{t}\|$  & stability difference at $k$-iteration on $t$-round \\
    \bottomrule  
  \end{tabular}  
\end{table} 

Then we introduce some important lemmas in our proofs.

\subsubsection{Important Lemmas}

\begin{lemma}\label{le:main_generation_error}
    (Lemma 3.11 in \cite{Hardt2016train}) We follow the definition in \cite{Hardt2016train,Zhou2021towards} to upper bound the uniform stability term after each communication round in DFL paradigm. Different from their vanilla calculations, DFL considers the finite-sum function on heterogeneous clients. Let non-negative objective $f$ is $L$-smooth and $L_G$-Lipschitz. After training $T$ rounds on $\mathcal{C}$ and $\widetilde{\mathcal{C}}$. our method generates two models $\bar{\mathbf{x}}^{T+1}$ and $\widetilde{\bar{\mathbf{x}}}^{T+1}$ respectively. For each data $z$ and every $t_0 \in \{1,2,3,\cdots,S\}$, we have:
    \begin{equation*}
        \mathbb{E}\|f({\bar{\mathbf{x}}}^{T+1};z)-f(\widetilde{\bar{\mathbf{x}}}^{T+1};z)\|\leq \frac{L_G}{m}\sum_{i=1}^m\mathbb{E}\left[\|\mathbf{x}_{i,K}^T - \widetilde{\mathbf{x}}_{i,K}^T\||\xi\right] + \frac{Ut_0}{S}
    \end{equation*}
    \begin{proof}
        Let $\xi = 1$ denote the event $\|\bar{\mathbf{x}}^{t_0} - \widetilde{\bar{\mathbf{x}}}^{t_0}\| = 0$ and $U = \sup_{\mathbf{x},z}\{ f ({\mathbf{x}}; z)\}$, we have:
        \begin{equation*}
            \begin{aligned}
                & \mathbf{E}\|f({\bar{\mathbf{x}}}^{T+1};z)-f(\widetilde{\bar{\mathbf{x}}}^{T+1};z)\| \\
                & = P(\{\xi\})\mathbb{E}\left[\|f({\bar{\mathbf{x}}}^{T+1};z)-f(\widetilde{\bar{\mathbf{x}}}^{T+1};z)\||\xi\right] + P(\{\xi^{c}\})\mathbb{E}\left[\|f({\bar{\mathbf{x}}}^{T+1};z)-f(\widetilde{\bar{\mathbf{x}}}^{T+1};z)\| |\xi^{c}\right]\\
                &\leq \mathbb{E}\left[\|f({\bar{\mathbf{x}}}^{T+1};z)-f(\widetilde{\bar{\mathbf{x}}}^{T+1};z)\||\xi\right] + P(\{\xi^{c}\})\sup_{\mathbf{x},z}\{ f ({\mathbf{x}}; z)\}\\
                &\leq L_G\mathbb{E}\left[\|{\bar{\mathbf{x}}}^{T+1}-\widetilde{\bar{\mathbf{x}}}^{T+1}\||\xi\right] + P(\{\xi^{c}\})U\\
                & =  L_G\mathbb{E}\left[\|\frac{1}{m}\sum_{i=1}^m(\mathbf{x}_{i,K}^T - \widetilde{\mathbf{x}}_{i,K}^T)\||\xi\right] + P(\{\xi^{c}\})U\\
                &\leq \frac{L_G}{m}\sum_{i=1}^m\mathbb{E}\left[\|\mathbf{x}_{i,K}^T - \widetilde{\mathbf{x}}_{i,K}^T\||\xi\right] + P(\{\xi^{c}\})U
            \end{aligned}
        \end{equation*}
        Before the $j^*$-th data on $i^*$-th client is sampled, the iterative states are identical on both $\mathcal{C}$ and $\widetilde{\mathcal{C}}$. Let $\widetilde{j}$ is the index of the first different sampling, if  $\widetilde{j} > t_0$, then $\xi = 1$ hold for $t_0$. Therefore, we have:
        \begin{equation*}
            P(\{\xi^{c}\}) =P(\{\xi = 0\}) =  P(\widetilde{j} \leq t_0) \leq \frac{t_0}{S}
        \end{equation*}
        We complete the proof.
    \end{proof}
\end{lemma}

\begin{lemma}\label{le:same}(Lemma 1.1 in \cite{Zhou2021towards})
    Different from their calculations, we prove similar inequalities on $f$ in the stochastic optimization. Under Assumption \ref{as:smoothness} and \ref{as:bounded_stochastic_gradient}, the local updates satisfy $\mathbf{x}_{i,k+1}^t = \mathbf{x}_{i,k}^t - \eta \mathbf{g}_{i,k}^t$ on $\mathcal{C}$ and $\widetilde{\mathbf{x}}_{i,k+1}^t = \widetilde{\mathbf{x}}_{i,k}^t - \eta \widetilde{\mathbf{g}}_{i,k}^t$ on $\widetilde{\mathcal{C}}$. If at $k$-th iteration on each round, we sample the \textbf{same} data in $\mathcal{C}$ and $\widetilde{\mathcal{C}}$, then we have:
    \begin{equation*}
        \mathbb{E}\|\mathbf{x}_{i,k+1}^{t}-\widetilde{\mathbf{x}}_{i,k+1}^{t}\| \leq (1+\eta L)\mathbb{E}\|\mathbf{x}_{i,k}^t-\widetilde{\mathbf{x}}_{i,k}^t\|+2\eta\sigma_l
    \end{equation*}
    \begin{proof}
    In each round $t$, by the triangle inequality and omitting the same data $z$, we have:
        \begin{equation*}
            \begin{aligned}
                &\mathbb{E}\|\mathbf{x}_{i,k+1}^{t}-\widetilde{\mathbf{x}}_{i,k+1}^{t}\| \\
                &=\mathbb{E}\|\mathbf{x}_{i,k}^t-\eta g_{i,k}^t-\widetilde{\mathbf{x}}_{i,k}^t-\eta\widetilde{g}_{i,k}^t\| \\
                &\leq\mathbb{E}\|w_{i,k}^t-\widetilde{\mathbf{x}}_{i,k}^t\|+\eta\mathbb{E}\|g_{i,k}^t-\widetilde{g}_{i,k}^t\| \\
                &\leq\mathbb{E}\|\mathbf{x}_{i,k}^t-\widetilde{\mathbf{x}}_{i,k}^t\|+\eta\mathbb{E}\|g_{i,k}^t-\nabla f_i(\mathbf{x}_{i,k}^t)\|+\eta\mathbb{E}\|\widetilde{g}_{i,k}^t-\nabla f_i(\widetilde{\mathbf{x}}_{i,k}^t)\| \\
                &\leq(1+\eta L)\mathbb{E}\|\mathbf{x}_{i,k}^t-\widetilde{\mathbf{x}}_{i,k}^t\|+2\eta\sigma_l
            \end{aligned}
        \end{equation*}
        The final inequality adopts assumptions of $\mathbb{E}\|g_{i,k}^t-\nabla f_i(w_{i,k}^t)\|\le\sqrt{\mathbb{E}\|g_{i,k}^t-\nabla f_i(w_{i,k}^t)\|^2}\le\sigma_l$. This completes the proof.
    \end{proof}
\end{lemma}

\begin{lemma}\label{le:different}(Lemma 1.2 in \cite{Zhou2021towards})
    Different from their calculations, we prove similar inequalities on $f$ in the stochastic optimization. Under Assumption \ref{as:smoothness}, \ref{as:L_G-lip} and \ref{as:bounded_stochastic_gradient}, the local updates satisfy $\mathbf{x}_{i,k+1}^t = \mathbf{x}_{i,k}^t - \eta \mathbf{g}_{i,k}^t$ on $\mathcal{C}$ and $\widetilde{\mathbf{x}}_{i,k+1}^t = \widetilde{\mathbf{x}}_{i,k}^t - \eta \widetilde{\mathbf{g}}_{i,k}^t$ on $\widetilde{\mathcal{C}}$. If at $k$-th iteration on each round, we sample the \textbf{different} data in $\mathcal{C}$ and $\widetilde{\mathcal{C}}$, then we have:
        \begin{equation*}
        \mathbb{E}\|\mathbf{x}_{i,k+1}^{t}-\widetilde{\mathbf{x}}_{i,k+1}^{t}\| \leq \mathbb{E}\|\mathbf{x}_{i,k}^t-\widetilde{\mathbf{x}}_{i,k}^t\|+2\eta(\sigma_l + L_G)
    \end{equation*}
    \begin{proof}
        In each round $t$, let by the triangle inequality and denoting the different data as $z$ and $\widetilde{z}$, we have:
        \begin{equation*}
            \begin{aligned}
                &\mathbb{E}\|\mathbf{x}_{i,k+1}^{t}-\widetilde{\mathbf{x}}_{i,k+1}^{t}\| \\
                &=\mathbb{E}\|\mathbf{x}_{i,k}^t-\eta g_{i,k}^t-\widetilde{\mathbf{x}}_{i,k}^t-\eta\widetilde{g}_{i,k}^t\| \\
                &\leq\mathbb{E}\|\mathbf{x}_{i,k}^t-\widetilde{\mathbf{x}}_{i,k}^t\|+\eta\mathbb{E}\|g_{i,k}^t-\widetilde{g}_{i,k}^t\| \\
                &= \mathbb{E}\|\mathbf{x}_{i,k}^t-\widetilde{\mathbf{x}}_{i,k}^t\| + \eta\mathbb{E}\|g_{i,k}^t-\nabla f_i(\mathbf{x}_{i,k}^t;z)-\widetilde{g}_{i,k}^t-\nabla f_i(\widetilde{\mathbf{x}}_{i,k}^t;\widetilde{z})+\nabla f_i(\mathbf{x}_{i,k}^t;z)-\nabla f_i(\widetilde{\mathbf{x}}_{i,k}^t;\widetilde{z})\|\\
                &\leq\mathbb{E}\|\mathbf{x}_{i,k}^t-\widetilde{\mathbf{x}}_{i,k}^t\|+2\eta\sigma_l+\eta\mathbb{E}\|\nabla f_i(\mathbf{x}_{i,k}^t;z)-\nabla f_i(\widetilde{\mathbf{x}}_{i,k}^t;\widetilde{z})\| \\
                &\leq\mathbb{E}\|\mathbf{x}_{i,k}^t-\widetilde{\mathbf{x}}_{i,k}^t\|+2\eta(\sigma_l+L_G).
            \end{aligned}
        \end{equation*}
        The final inequality adopts the $L_G$-Lipschitz. This completes the proof.
    \end{proof}
\end{lemma}

\begin{lemma}\label{le:ave}
    Given the stepsize $ \beta \leq \frac{1-\psi}{4\sqrt{m} + 1-\psi}$, we have following bound:
    \begin{equation*}
        \mathbf{E}\|\mathbf{x}_{i}^t - \bar{\mathbf{x}}^t\| \leq \frac{1+\beta}{\alpha(1-\beta)}K(\sigma_l + L_G)\sum_{j=0}^{t-1}\eta_j\psi^{t-1-j}
    \end{equation*}
    Where $\alpha = 1 - \frac{4\sqrt{m}\beta}{(1-\psi)(1-\beta)}$.
    \begin{proof}
        Following [Lemma 4, \cite{Sun2022Decentralized}], 
        we denote ${\bf Z}^{t}:=\begin{bmatrix}
        {\bf z}_1^{t},  {\mathbf z}_2^{t},
        \ldots,
        {\bf z}_m^{t}
        \end{bmatrix}^{\top}\in\mathbb{R}^{m\times d}$.
        With these notation, we have
        \begin{align}\label{xtglobal}
        {\bf X}^{t+1}={\bf W}{\bf Z}^{t}={\bf W}{\bf X}^{t}-{\bf \zeta}^t,
        \end{align}
        where ${\bf \zeta}^t:={\bf W}{\bf X}^{t}-{\bf W}{\bf Z}^{t}$. Following [Lemma 8, \cite{Sun2022Stability}], we have:
        \begin{equation}\label{recu}
            \begin{aligned}
                \mathbf{E} \|(\mathbb{I}-\mathbf{P})\mathbf{X}^{t+1}\| \leq \psi \mathbf{E} \|(\mathbb{I}-\mathbf{P})\mathbf{X}^{t}\| + 2\mathbf{E} \| \zeta^t\|
            \end{aligned}
        \end{equation}
        Assuming $\mathbf{E} \|\mathbf{x}_{i}^t - \bar{\mathbf{x}}^t\| \leq D$ , it means that $\mathbf{E} \|(\mathbb{I}-\mathbf{P})\mathbf{X}^{t}\| \leq \sqrt{m}D$. Next, we will bound $\mathbf{E} \| \mathbf{x}_{i,k}^t - \mathbf{x}_i^t\|$. According to the equation $\mathbf{x}_{i,k}^t - \mathbf{x}_i^t = \beta (\mathbf{x}_i^t - \mathbf{x}_{i,K}^{t-1} ) - \eta \sum_{j=0}^{k-1}\mathbf{g}_{i,j}^t$ and Assumption \ref{as:L_G-lip}, we have:
        \begin{equation*}
            \begin{aligned}
                \mathbf{E} \| \mathbf{x}_{i,k}^t - \mathbf{x}_i^t\| \leq \beta\mathbf{E} \|\mathbf{x}_i^t - \mathbf{x}_{i,K}^{t-1}\| + K\eta(\sigma_l +L_G)
            \end{aligned}
        \end{equation*}
        Next, we need to bound $\mathbf{E} \| \mathbf{x}_i^t - \mathbf{x}_{i,K}^{t-1} \|$.
        According to (\ref{18}), we have:
        \begin{equation*}
            \begin{aligned}
                \mathbf{E}\|\mathbf{x}_i^{t+1} - \mathbf{x}_{i,K}^{t}\| &\leq \beta \mathbf{E}\| \mathbf{x}_i^t - \mathbf{x}_{i,K}^{t-1} \| + \mathbf{E} \|\mathbf{x}_i^{t+1} - \mathbf{x}_i^{t+1}\| + \mathbf{E} \|\eta \sum_{k=0}^{K-1} \mathbf{g}_{i,k}^t\|\\
                & \leq \beta \mathbf{E} \| \mathbf{x}_i^t - \mathbf{x}_{i,K}^{t-1} \| + \mathbf{E} \|\mathbf{x}_i^{t+1} - \bar{\mathbf{x}}^{t+1}\| + \mathbf{E} \|\mathbf{x}_i^{t} - \bar{\mathbf{x}}^{t}\|+ \mathbf{E} \|\bar{\mathbf{x}}^{t+1} - \bar{\mathbf{x}}^{t}\| + \mathbf{E} \|\eta \sum_{k=0}^{K-1} \mathbf{g}_{i,k}^t\|\\
                &\overset{(a)}\leq \beta \mathbf{E} \| \mathbf{x}_i^t - \mathbf{x}_{i,K}^{t-1} \| + 2D + 2K\eta(\sigma_l + L_G)\\
                &\leq \frac{2D + 2K\eta(\sigma_l + L_G)}{1-\beta}
            \end{aligned}
        \end{equation*}
        Where (a) uses  $\mathbf{E}\|\mathbf{x}_{i}^t - \bar{\mathbf{x}}^t\| \leq D$,  $\mathbf{E}\|\eta \sum_{k=0}^{K-1} \mathbf{g}_{i,k}^t\| \leq K\eta(\sigma_l + L_G)$ and $\mathbf{E} \|\bar{\mathbf{x}}^{t+1} - \bar{\mathbf{x}}^{t}\| = \mathbf{E}\| \frac{1}{m}\sum_{i=1}^m\sum_{k=0}^{K-1} \eta\mathbf{g}_{i,k}^t\| \leq K\eta(\sigma_l + L_G)$. Then we get 
        \begin{equation*}
            \begin{aligned}
                \mathbf{E}\| \mathbf{x}_{i,k}^t - \mathbf{x}_i^t\| \leq \frac{2\beta}{1-\beta}\left( D + K\eta(\sigma_l + L_G)\right) + K\eta(\sigma_l +L_G) = \frac{2\beta}{1-\beta}D + \left( \frac{1+\beta}{1-\beta}\right)K\eta (\sigma_l + L_G)
            \end{aligned}
        \end{equation*}
        This implies:
        \begin{equation*}
            \mathbf{E}\|\zeta^t\| \leq \sqrt{m}\left( \frac{2\beta}{1-\beta}D + \left( \frac{1+\beta}{1-\beta}\right)K\eta (\sigma_l + L_G) \right)
        \end{equation*}
        According to (\ref{recu}), we have:
        \begin{equation} \label{24}
            \begin{aligned}
                \mathbf{E}\|(\mathbb{I}-\mathbf{P})\mathbf{X}^{t}\| \leq 2\sqrt{m}\left(\frac{2\beta}{(1-\psi)(1-\beta)}D +\frac{1+\beta}{1-\beta}K(\sigma_l + L_G)\sum_{j=0}^{t-1}\eta_j\psi^{t-1-j}\right)
            \end{aligned}
        \end{equation}
        Letting the term on the right-hand side of (\ref{24}) be denoted as $D$, we obtain that, when $\beta \leq \frac{1-\psi}{4\sqrt{m} + 1-\psi}$ and let $\alpha = 1 - \frac{4\sqrt{m}\beta}{(1-\psi)(1-\beta)}$, we have 
        \begin{equation*}
            \mathbf{E}\|\mathbf{x}_{i}^t - \bar{\mathbf{x}}^t\| \leq \frac{1+\beta}{\alpha(1-\beta)}K(\sigma_l + L_G)\sum_{j=0}^{t-1}\eta_j\psi^{t-1-j}
        \end{equation*}
        We have completed the proof.
    \end{proof}
\end{lemma}

\begin{lemma}\label{le:inequation}
    (Lemma 5 in \cite{Sun2022Stability}) For any $0 < \psi < 1$ and $t\in\mathbb{Z}^{+}$, it holds
    \begin{equation*}
        \sum_{j=0}^{t-1}\frac{\psi^{t-1-j}}{j+1}\leq\frac{C_\lambda}t
    \end{equation*}
    with $ \left.C_\lambda:=\left\{\begin{array}{cc}\ln\frac{1}{\lambda}\frac{\lambda^{\ln\frac{1}{\lambda}}}{\lambda}+\frac{\ln^2\frac{1}{\lambda}}{16\lambda}\lambda^{\frac{\ln\frac{1}{\lambda}}{8}}+\frac{2}{\lambda\ln\frac{1}{\lambda}}\lambda\neq0,\\0,\lambda=0.\end{array}\right.\right.$
\end{lemma}

\subsubsection{Bounded Uniform Stability}
According to Lemma \ref{le:main_generation_error}, we firstly bound the recursive stability on $k$ in one round. If the sampled data is the same, we can adopt Lemma \ref{le:same}. Otherwise, we adopt Lemma \ref{le:different}. Thus we can bound the first term in Lemma \ref{le:main_generation_error} as:
\begin{equation*}
    \begin{aligned}
        \delta_{k+1}^t &= \frac{1}{m}\sum_{i=1}^m\mathbb{E}\left[\|\mathbf{x}_{i,k+1}^t - \widetilde{\mathbf{x}}_{i,k+1}^t\||\xi\right]\\
        & = P(z)\frac{1}{m}\sum_{i=1}^m\mathbb{E}\left[\|\mathbf{x}_{i,k+1}^t - \widetilde{\mathbf{x}}_{i,k+1}^t\||\xi ,z\right] + P(\widetilde{z})\frac{1}{m}\sum_{i=1}^m\mathbb{E}\left[\|\mathbf{x}_{i,k+1}^t - \widetilde{\mathbf{x}}_{i,k+1}^t\||\xi ,\widetilde{z}\right] \\
        &\leq \frac{S-1}{mS}\sum_{i=1}^m\left((1+\eta L)\mathbb{E}\left[\|\mathbf{x}_{i,k}^t-\widetilde{\mathbf{x}}_{i,k}^t\| | \xi\right] + 2\eta\sigma_l\right) + \frac{1}{mS}\sum_{i=1}^m\left( \mathbb{E}\left[\|\mathbf{x}_{i,k}^t-\widetilde{\mathbf{x}}_{i,k}^t\|| \xi\right] + 2\eta(\sigma_l+L_G)\right)\\
        & \leq (1+\eta L)\delta_{k}^t + \frac{2\eta L_G}S+2\eta\sigma_l
    \end{aligned}
\end{equation*}

Balancing the LHS and RHS, we have the following recursive formulation:
\begin{equation*}
    \delta_{k+1}^t + \frac{2(L_G+S\sigma_l)}{SL} \leq (1+\eta L)\left( \delta_{k}^t + \frac{2(L_G+S\sigma_l)}{SL} \right)
\end{equation*}

Therefore, in one single communication round, by generally defining learning rate $\eta = \eta_{k}^t$:
\begin{equation*}
    \delta_{K}^t + \frac{2(L_G+S\sigma_l)}{SL} \leq \left(\prod_{k=0}^{K-1}(1+\eta_k^tL)\right)\left( \delta_{0}^t + \frac{2(L_G+S\sigma_l)}{SL} \right)
\end{equation*}

The next important relationship is to measure the $\delta_{K}^{t-1}$ and $\delta_{0}^t$. According to the update rule $\mathbf{x}_{i,0}^t = 
\mathbf{x}_{i}^t + \beta(\mathbf{x}_{i}^t - \mathbf{x}_{i,K}^
{t-1})$, we have the difference follows:
\begin{equation*}
\begin{aligned}
    \mathbf{x}_{i,0}^t - \widetilde{\mathbf{x}}_{i,0}^t &= \mathbf{x}_{i}^t - \widetilde{\mathbf{x}}_{i}^t + \beta(\mathbf{x}_{i}^t - \mathbf{x}_{i,K}^
    {t-1}) -\beta(\widetilde{\mathbf{x}}_{i}^t - \widetilde{\mathbf{x}}_{i,K}^
    {t-1})\\
    & = (1+\beta)(\mathbf{x}_{i}^t - \widetilde{\mathbf{x}}_{i}^t) - \beta(\mathbf{x}_{i,K}^
    {t-1} - \widetilde{\mathbf{x}}_{i,K}^{t-1})
\end{aligned}
\end{equation*}
By taking the expectation on the $l_2$ norm, we have:
\begin{equation*}
\begin{aligned}
    \delta_0^t = \frac{1}{m}\sum_{i=1}^m\mathbb{E}\| \mathbf{x}_{i,0}^t - \widetilde{\mathbf{x}}_{i,0}^t \| 
    &\leq \frac{1+\beta}{m}\sum_{i=1}^m\mathbb{E}\| \mathbf{x}_{i}^t - \widetilde{\mathbf{x}}_{i}^t\| + \frac{\beta}{m}\sum_{i=1}^m\mathbb{E}\|\mathbf{x}_{i,K}^
    {t-1} - \widetilde{\mathbf{x}}_{i,K}^{t-1}\|\\
    &\leq (1+\beta)\underbrace{\frac{1}{m}\sum_{i=1}^m\mathbb{E}\| \mathbf{x}_{i}^t - \widetilde{\mathbf{x}}_{i}^t\|}_{R_1} + \beta \delta_K^{t-1}
\end{aligned}
\end{equation*}
Next, we bound R1.
\begin{equation*}
    \begin{aligned}
        R_1 &= \frac{1}{m}\sum_{i=1}^m\mathbb{E}\| \mathbf{x}_{i}^t - \widetilde{\mathbf{x}}_{i}^t\| \\
        &\leq \frac{1}{m}\sum_{i=1}^m\mathbb{E}\| \mathbf{x}_{i}^t - \bar{\mathbf{x}}^t\| + \frac{1}{m}\sum_{i=1}^m\mathbb{E}\| \widetilde{\mathbf{x}}_{i}^t - \widetilde{\bar{\mathbf{x}}}^t\| + \frac{1}{m}\sum_{i=1}^m\mathbb{E}\| \widetilde{\bar{\mathbf{x}}}^t - \bar{\mathbf{x}}^t\| \\
        & = \frac{1}{m}\sum_{i=1}^m\mathbb{E}\| \mathbf{x}_{i}^t - \bar{\mathbf{x}}^t\| + \frac{1}{m}\sum_{i=1}^m\mathbb{E}\| \widetilde{\mathbf{x}}_{i}^t - \widetilde{\bar{\mathbf{x}}}^t\| + \mathbb{E}\| \frac{1}{m}\sum_{i=1}^m(\mathbf{x}_{i,K}^{t-1} - \widetilde{\mathbf{x}}_{i,K}^{t-1})\| \\
        &\leq \frac{1}{m}\sum_{i=1}^m\mathbb{E}\| \mathbf{x}_{i}^t - \bar{\mathbf{x}}^t\| + \frac{1}{m}\sum_{i=1}^m\mathbb{E}\| \widetilde{\mathbf{x}}_{i}^t - \widetilde{\bar{\mathbf{x}}}^t\| +  \frac{1}{m}\sum_{i=1}^m\mathbb{E}\|\mathbf{x}_{i,K}^{t-1} - \widetilde{\mathbf{x}}_{i,K}^{t-1}\|\\
        &\leq \frac{1}{m}\sum_{i=1}^m\mathbb{E}\| \mathbf{x}_{i}^t - \bar{\mathbf{x}}^t\| + \frac{1}{m}\sum_{i=1}^m\mathbb{E}\| \widetilde{\mathbf{x}}_{i}^t - \widetilde{\bar{\mathbf{x}}}^t\| + \delta_K^{t-1}\\
        & \leq 2D_t + \delta_K^{t-1}
    \end{aligned}
\end{equation*}
Where we used Lemma \ref{le:ave} and Lemma \ref{le:inequation} to establish the final inequality and let the learning rate be the same selection as it in the optimization of $\mathcal{O}(\frac{1}{t}) = \frac{c}{t}$, then we get $D_t = \frac{1}{\alpha}K(\sigma_l + L_G)\frac{C_\lambda}{t}$. Finally we get
\begin{equation*}
    \delta_0^t \leq  (1+2\beta)\delta_K^{t-1} + 2(1+\beta)D_t
\end{equation*}
By denoting $ \phi(t)=\prod_{k=0}^{K-1}(1+\eta_k^tL) $be the combination of learning rate $\eta_k^t$, we can provide an upper bound of the recursive formulation as:
\begin{equation*}
    \delta_{K}^t + \frac{2(L_G+S\sigma_l)}{SL} \leq \left(\prod_{k=0}^{K-1}(1+\eta_k^tL)\right)\left( \delta_{0}^t + \frac{2(L_G+S\sigma_l)}{SL} \right) \leq \phi(t)\left( (1+2\beta)\delta_K^{t-1} + 2(1+\beta)D_t + \frac{2(L_G+S\sigma_l)}{SL}\right)
\end{equation*}
To balance the constant part, assuming the learning rate is decayed by communication round $t$ which indicates $\phi(t) \leq \phi(t-1)$ and let $1+2\beta \leq \frac{\phi(t-1)}{\phi(t)}$ be the upper bound and small learning rates \(\eta_k^t\) ensure that \(\frac{4\beta(L_G+S\sigma_l)}{SL} \geq 2(1+\beta)D_t\), then we have the following recursive formulation:
\begin{equation*}
    \delta_{K}^t + \frac{\phi(t) - 1}{(1+2\beta)\phi(t) - 1}A +   \frac{\phi(t)}{(1+2\beta)\phi(t) - 1}C_t \leq  \phi(t-1)\left( \delta_K^{t-1} + \frac{\phi(t-1) - 1}{(1+2\beta)\phi(t-1) - 1}A +   \frac{\phi(t-1)}{(1+2\beta)\phi(t-1) - 1}C_{t-1}\right)
\end{equation*}
Where $A = \frac{2(L_G+S\sigma_l)}{SL}, C_t = 2(1+\beta)D_t = \frac{2(1+\beta)}{\alpha}K(\sigma_l + L_G)\frac{C_\lambda}{t}$, Unrolling from $t_0 - 1$ to $T$, we have:
\begin{equation*}
\begin{aligned}
    \delta_K^{T+1} &\leq \left(\prod_{\tau = t_0-1}^T \phi(\tau)\right)\left( \delta_K^{t_0 -1} + \frac{\phi(t_0 -1) - 1}{(1+2\beta)\phi(t_0 - 1) - 1}A +   \frac{\phi(t_0 - 1)}{(1+2\beta)\phi(t_0 - 1) - 1}C_{t_0 -1}\right) \\
    &\quad - \left(\frac{\phi(T+1) - 1}{(1+2\beta)\phi(T+1) - 1}A + \frac{\phi(T+1)}{(1+2\beta)\phi(T+1) - 1}C_{t_0 -1}\right)\\
    &\leq \left(\prod_{\tau = t_0-1}^T \phi(\tau)\right)\left( \frac{\phi(t_0 -1) - 1}{(1+2\beta)\phi(t_0 - 1) - 1}A +   \frac{\phi(t_0 - 1)}{(1+2\beta)\phi(t_0 - 1) - 1}C_{t_0 -1}\right)\\
    &\leq \left(\prod_{\tau = t_0-1}^T\prod_{k=0}^{K-1}(1+\eta_{k}^{\tau}L)\right)\left( \frac{A + C_{t_0 -1}}{(1+2\beta)} \right)\\
    &\leq \exp{\left(\sum_{\tau = t_0-1}^T\sum_{k=0}^{K-1}\eta_{k}^{\tau}L\right)}\left( \frac{2(L_G+S\sigma_l)}{(1+2\beta)SL} +\frac{2(1+\beta)K(\sigma_l + L_G)}{\alpha(1+2\beta)}\frac{C_\lambda}{t_0 }\right)
\end{aligned}
\end{equation*}
According to the preceding selection of the learning rate, \(\mathcal{O}(\frac{1}{t}) = \frac{c}{t}\), we have:
\begin{equation*}
\begin{aligned}
    \delta_{K}^T &\leq  \exp{\left(\sum_{\tau = t_0-1}^{T-1}\sum_{k=0}^{K-1}\eta_{k}^{\tau}L\right)}\left( \frac{2(L_G+S\sigma_l)}{(1+2\beta)SL} +\frac{2(1+\beta)K(\sigma_l + L_G)}{\alpha(1+2\beta)}\frac{C_\lambda}{t_0}\right)\\
    &\leq \exp{\left(\sum_{\tau = t_0-1}^{T-1}\frac{cKL}{\tau }\right)}\left( \frac{2(L_G+S\sigma_l)}{(1+2\beta)SL} +\frac{2(1+\beta)K(\sigma_l + L_G)}{\alpha(1+2\beta)}\frac{C_\lambda}{t_0}\right)\\
    &\leq \exp\left(\int_{\tau=t_0}^T\frac{cKL}\tau d\tau\right)\left( \frac{2(L_G+S\sigma_l)}{(1+2\beta)SL} +\frac{2(1+\beta)K(\sigma_l + L_G)}{\alpha(1+2\beta)}\frac{C_\lambda}{t_0}\right)\\
    &\leq \left(\frac{T}{t_{0}}\right)^{cKL}\left( \frac{2(L_G+S\sigma_l)}{(1+2\beta)SL} +\frac{2(1+\beta)K(\sigma_l + L_G)}{\alpha(1+2\beta)}\frac{C_\lambda}{t_0}\right)
\end{aligned}
\end{equation*}
To summarize the above inequalities and the Lemma \ref{le:main_generation_error}, we have:
\begin{equation*}
    \begin{aligned}
        &\mathbb{E}\|f({\bar{\mathbf{x}}}^{T+1};z)-f(\widetilde{\bar{\mathbf{x}}}^{T+1};z)\| 
        \leq L_G \delta_K^T + \frac{Ut_0}{S} \\
        &\leq L_G \left(\frac{T}{t_{0}}\right)^{cKL}\left( \frac{2(L_G+S\sigma_l)}{(1+2\beta)SL} +\frac{2(1+\beta)K(\sigma_l + L_G)}{\alpha(1+2\beta)}\frac{C_\lambda}{t_0}\right) + \frac{Ut_0}{S}
    \end{aligned}
\end{equation*}
Setting $t_0 = T^{\frac{cKL}{1+cKL}}\left( (\frac{2L_G(L_G+S\sigma_l)}{(1+2\beta)SL} +\frac{2L_G(1+\beta)K(\sigma_l + L_G)}{\alpha(1+2\beta)}C_\lambda)\frac{ScKL}{U}\right)^{\frac{1}{1+cKL}}$, We then have
\begin{equation*}
    \begin{aligned}
        \mathbb{E}\|f({\bar{\mathbf{x}}}^{T+1};z)-f(\widetilde{\bar{\mathbf{x}}}^{T+1};z)\| 
        \leq 2T^{\frac{cKL}{1+cKL}}\left( \frac{2L_G(L_G+S\sigma_l)}{(1+2\beta)SL} +\frac{2L_G(1+\beta)K(\sigma_l + L_G)}{\alpha(1+2\beta)}C_\lambda\right)^{\frac{1}{1+cKL}}\left( \frac{U}{S}\right)^{\frac{cKL}{1+cKL}}
    \end{aligned}
\end{equation*}